\documentclass{article}

\usepackage{iclr2027_conference,times}
\iclrfinalcopy

\usepackage{amsmath}
\usepackage{amssymb}
\usepackage{booktabs}
\usepackage{longtable}
\usepackage{colortbl}
\usepackage{graphicx}
\usepackage{wrapfig}
\usepackage{microtype}
\usepackage{multirow}
\usepackage{tikz}
\usepackage{xcolor}
\usepackage{xspace}
\usepackage{url}
\usepackage{hyperref}

\IfFileExists{pcrr7t.tfm}{}{}

\definecolor{softgreen}{rgb}{0.03,0.60,0.27}
\newcommand{\systemname}{RTL-BenchLS\xspace}
\newcommand{\filledcircled}[1]{%
  \tikz[baseline=(char.base)]{%
    \node[circle,fill=black,text=white,inner sep=0.6pt,
      font=\scriptsize\bfseries] (char) {#1};}}

\makeatletter
\newenvironment{figureblock}
  {\par\addvspace{\intextsep}\noindent\def\@captype{figure}}
  {\par\addvspace{\intextsep}}
\newenvironment{tableblock}
  {\par\addvspace{\intextsep}\noindent\def\@captype{table}}
  {\par\addvspace{\intextsep}}
\makeatother

\newcommand{\taskname}[1]{\csname taskname@#1\endcsname}
\expandafter\def\csname taskname@1.1\endcsname{Direct Module-level Generation}
\expandafter\def\csname taskname@1.2\endcsname{Direct Project-level Generation}
\expandafter\def\csname taskname@1.3\endcsname{Round-trip Module-level Generation}
\expandafter\def\csname taskname@1.4\endcsname{Round-trip Project-level Generation}
\expandafter\def\csname taskname@2.1\endcsname{Direct Module-level Completion}
\expandafter\def\csname taskname@2.2\endcsname{Direct Project-level Completion}
\expandafter\def\csname taskname@2.3\endcsname{Round-trip Module-level Completion}
\expandafter\def\csname taskname@2.4\endcsname{Round-trip Project-level Completion}
\expandafter\def\csname taskname@3.1\endcsname{Direct Module-level Debugging}
\expandafter\def\csname taskname@3.2\endcsname{Direct Project-level Debugging}
\expandafter\def\csname taskname@3.2a\endcsname{Direct Project-level Debugging (nonfunctional issue)}
\expandafter\def\csname taskname@3.2b\endcsname{Direct Project-level Debugging (functional issue)}
\expandafter\def\csname taskname@3.2c\endcsname{Direct Project-level Debugging (mutant)}
\expandafter\def\csname taskname@3.3\endcsname{Round-trip Module-level Debugging}
\expandafter\def\csname taskname@3.4\endcsname{Round-trip Project-level Debugging}

\title{RTL-BenchLS: A Large-Scale Benchmark for Agentic RTL Reasoning and Implementation}

\author{\textbf{Jing Wang, Shang Liu, Wenji Fang, Yuchao Wu, Yugao Zhu, Zhiyao Xie\thanks{Corresponding author.}} \\[1ex]
The Hong Kong University of Science and Technology \\[0.5ex]
\texttt{jwangjw@connect.ust.hk, eezhiyao@ust.hk}}

\hypersetup{
  pdftitle={RTL-BenchLS: A Large-Scale Benchmark for Agentic RTL Reasoning and Implementation},
  pdfauthor={Jing Wang, Shang Liu, Wenji Fang, Yuchao Wu, Yugao Zhu, Zhiyao Xie}
}

\begin{document}

\maketitle
\fancyhead{}
\renewcommand{\headrulewidth}{0pt}

\begin{abstract}

Agentic frameworks are advancing rapidly, driving growing interest in agents for automated hardware design. Implementing hardware in RTL (i.e., RTL generation) is a key task in VLSI circuit design. However, \emph{existing RTL benchmarks are too easy for cutting-edge agentic frameworks.} Specifically, we observe three gaps in the RTL benchmarks. \emph{(1) Limited task scope.} Benchmarks are limited in module-level scope. Such module-level scope is insufficient to evaluate agentic frameworks. The EDA community urgently requires a qualified project-level RTL benchmark. \emph{(2) Narrow task type.} Most benchmarks only focus on a narrow task scope. A single narrow task type is not enough to thoroughly evaluate agents on complicated RTL tasks. \emph{(3) Small benchmark scale and design size.} Existing benchmarks are usually small in scale with small design size. To address these limitations, we introduce \textbf{RTL-BenchLS}, a large-scale benchmark of 17K+ cases with 12 tasks. We contribute in three aspects. \emph{(1) New task scope.} We expand the scope from \emph{module level} to \emph{project level} with realistic project-level designs. Experiments show that our module-level cases are sufficiently challenging, and project-level tasks expose obvious limitations of current agentic frameworks. \emph{(2) Expanded task type.} We cover three important types to evaluate agents' abilities from different perspectives, including generation, completion, and debugging.  \emph{(3) Enlarged benchmark scale and design size.} We collect a significantly larger scale of cases covering larger design sizes. To further increase the benchmark scale, we define a new task \emph{protocol} named the \emph{round-trip} protocol, which eliminates the need to manually craft qualified specifications and testbenches. As a result, we can significantly expand the benchmark scale using collected RTL designs. The \emph{round-trip} tasks are also significantly challenging for the current LLMs. This new benchmark, at unprecedented scale, will help evaluate academic work and industrial solutions and improve LLM/agent solutions for hardware RTL design. Our benchmark will be released at \url{https://github.com/hkust-zhiyao/RTL-BenchLS}.\footnote{This is RTL-BenchLS v1.0. We will continue to update the benchmark.}
\end{abstract}

\section{Introduction}
\label{sec:intro}

The rapid advancement of agentic frameworks has spurred growing interest in agents for automated hardware design~\citep{pearce2020dave,thakur2024verigen,chang2023chipgpt,thakur2023autochip,liu2024rtlcoder,cui2024origen,ho2024verilogcoder,pei2024betterv,auto-v-coder,zhao2024mage}. Industry is also placing increasing emphasis on automated RTL implementation. For example, Cadence recently introduced an RTL Generation Agent within its ChipStack AI Super Agent.\footnote{\href{https://newsroom.cadence.com/press-releases/press-release-details/2026/Cadence-Expands-ChipStack-AI-Super-Agent-with-a-New-Agent-for-RTL-Generation-and-Early-PPA-Optimization/default.aspx}{Cadence, ``RTL Generation and Early PPA Optimization,'' 2026.}} Implementing hardware in RTL (i.e., RTL generation) is a key task in VLSI circuit design. All VLSI circuits, including CPUs, GPUs, and AI accelerators, are implemented in RTL code (e.g., Verilog or VHDL code). However, \emph{existing RTL benchmarks are too easy for cutting-edge agentic frameworks}~\citep{kang2024fveval,huang2025rtlspec,zhu2025codevr1,pinckney2025cvdp,liu2025deeprtl}. For example, MAGE~\citep{zhao2024mage} and VerilogCoder~\citep{ho2024verilogcoder} achieve functional pass rates above 90\% on the RTL benchmark VerilogEval~\citep{liu2023verilogeval, pinckney2024revisiting}. Additionally, some failures are caused by flaws in the cases themselves~\citep{wang2026rtl}. As a result, current benchmarks leave limited room to distinguish and further improve agentic frameworks.

We observe three gaps in existing benchmarks on LLMs for RTL design. \textbf{(1) Limited task scope.} Widely used benchmarks such as VerilogEval~\citep{liu2023verilogeval,pinckney2024revisiting} and RTLLM~\citep{lu2024rtllm,liu2025openllmrtl} primarily assess RTL generation on module-level scope. Such module-level designs do not represent realistic industrial scenarios. As a result, the benchmarks are insufficient to evaluate cutting-edge agents on more realistic tasks. The EDA community urgently needs a qualified RTL benchmark at the \emph{project level}. \textbf{(2) Narrow task type.} Existing benchmarks mainly focus on a single type of task, such as generation~\citep{lu2024rtllm, liu2023verilogeval, pinckney2024revisiting} or completion~\citep{allam2024rtlrepo, jin2025realbench} task. A single narrow task type is not enough to thoroughly evaluate agents on complicated RTL tasks. \textbf{(3) Small benchmark scale and design size.} Existing benchmarks are small in scale and usually contain small designs. For example, the most representative benchmarks, VerilogEval~\citep{liu2023verilogeval} and RTLLM~2.0~\citep{lu2024rtllm}, contain 156 and 50 designs, respectively. CVDP provides 783 cases with larger designs, but it has fewer than 200 cases per task. 

\begin{figure}[!t]
\vspace{-.2in}
  \centering
  \includegraphics[width=\textwidth]{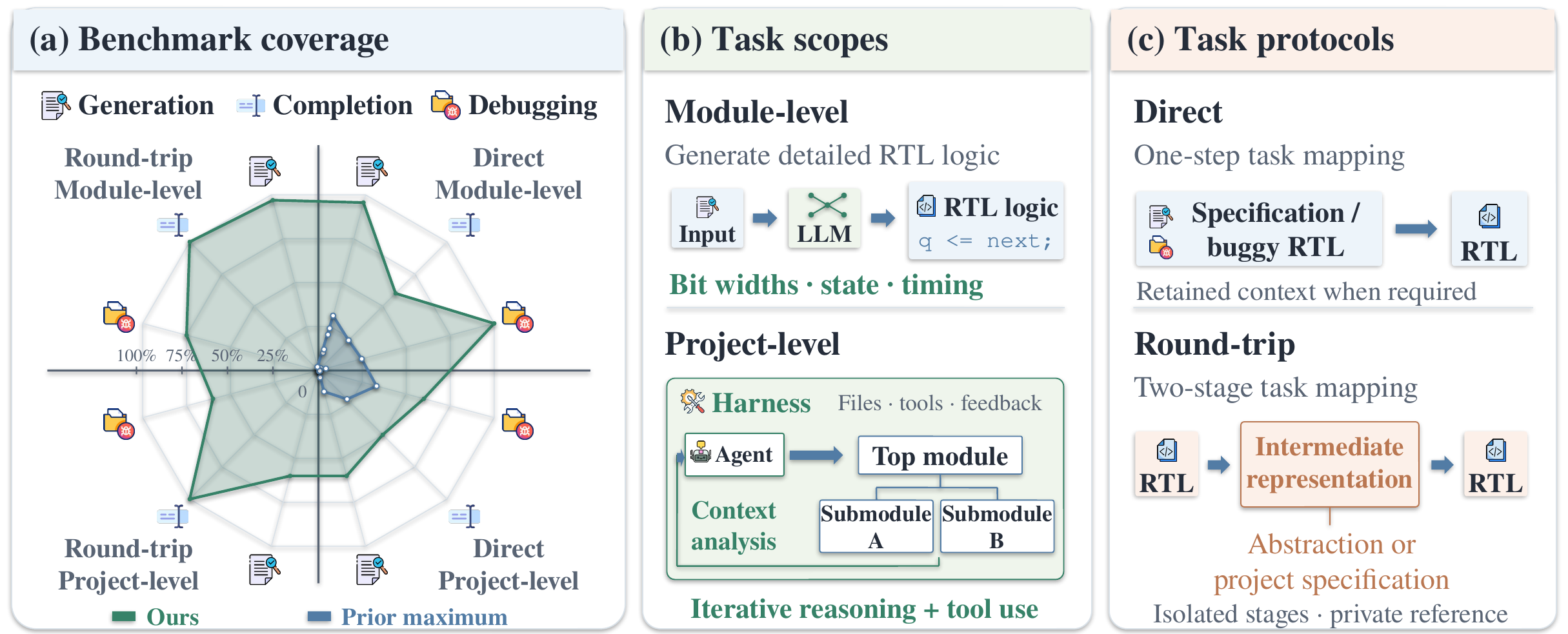}
  \vspace{-.1in}
  \caption{Overview of RTL-BenchLS. \emph{(a) Benchmark coverage.} Our benchmark provides significantly larger scale and more tasks than existing benchmarks. \emph{(b) Task scope.} In addition to traditional module-level tasks, we introduce project-level RTL tasks. (c) \emph{Task protocols.} In additional to the traditional \emph{direct} task, we propose a new \emph{round-trip} task.}
  \vspace{-.2in}
  \label{fig:intro-overview}
\end{figure}

To address the limitations, we introduce \textbf{RTL-BenchLS}, a large-scale RTL benchmark with 17K+ cases and 12 tasks. {Specifically, the 17K+ cases originate from 9000+ RTL deisgns. We construct 16K+ \emph{module-level} task cases based on 9K+ RTL modules, and construct 2000+ \emph{project-level} task cases from 70+ RTL projects.} {In the following content, we will introduce our contributions from three aspects,} each targeting a specific limitation.

{\textbf{(1) New task scope.} We expand the task scope from module level to project level with realistic project-level designs. Our benchmark includes two task scopes. Figure~\ref{fig:intro-overview}(b) depicts the two scopes and the distinct evaluation focus. The traditional \emph{module-level} task scopes evaluate LLMs on detailed RTL logic within one module. We introduce the \emph{project-level} scope, which requires an agent to understand project context, call tools, and finish the task iteratively.} Experiments show that our cases at the existing module-level scope are sufficiently challenging, and project-level scope tasks help expose limitations of agentic frameworks.

{\textbf{(2) Expanded task type.} To evaluate agents' ability thoroughly from different perspectives, we cover three important types, including generation, completion, and debugging. Specifically, \emph{generation} creates RTL from a specification. \emph{Completion} fills missing logic or modules using the specification and surrounding context. \emph{Debugging} locates and corrects faulty RTL or reported issues. These task types require different reasoning and generation abilities of agents.}

{\textbf{(3) A new task protocol for scaling up.}  To further increase the benchmark scale, we propose a new task \textbf{protocol} called the \textbf{round-trip} task. Figure~\ref{fig:intro-overview}(c) depicts the scenario for two task protocols. We define traditional RTL tasks as the \emph{direct} protocol, where the task input is a supplied specification, RTL context, or issue report. In the \emph{Round-trip} protocol, the agent must perform a two-stage task. In the first stage, the agent generates an intermediate representation of RTL code. In the second step, the agent takes the intermediate representation as input to recover or debug the RTL code. The \emph{round-trip} protocol eliminates the need to manually craft qualified specifications and testbenchs. As a result, the benchmark scale can be significantly expanded from collected RTL designs. Experiments show that tasks in the \emph{round-trip protocol} are significantly more challenging than traditional tasks with \emph{direct protocol}.}

\textbf{Enlarged benchmark scale and design size.} We collect a significantly more cases covering larger design sizes.
Figure~\ref{fig:intro-overview}(a) provides an overview of the benchmark. We contribute more cases on existing tasks and expand to new tasks. Specifically, we construct 12 tasks based on the three dimensions: two task \emph{scopes}, three task \emph{types}, and two task \emph{protocols}. 
This new benchmark, at unprecedented scale, will help evaluate academic work and industrial solutions and improve LLM/agent solutions for hardware RTL design.

\section{Related Work}
\label{sec:related}

Table~\ref{tab:benchmark-comparison} compares \emph{RTL-BenchLS} with representative RTL benchmarks on different tasks. In the table, tasks are organized based on three dimensions: \emph{protocol}, \emph{scope}, and \emph{type}. The numbers show the number of cases for each task. \emph{RTL-BenchLS} contributes more tasks with larger scale and design size. Figure~\ref{fig:design-statistics} compares \emph{RTL-BenchLS} with representative benchmarks. Our benchmark contributes more cases across the design sizes. Next, we discuss the benchmark comparison at \emph{module-level scope} and \emph{project-level scope}.

\textbf{Module-level scope.} The most widely used RTL generation benchmarks primarily evaluate tasks at \emph{module-level scope} and they are mainly limited in benchmark scale and design size. Specifically, VerilogEval~\citep{liu2023verilogeval,pinckney2024revisiting} and RTLLM 2.0~\citep{liu2025openllmrtl} contain 156 and 50 designs, respectively, and are mostly simple. CVDP contributes larger designs with more tasks, but remains limited in benchmark scale, with fewer than 200 cases per task. \emph{RTL-BenchLS} provides more and larger cases across the tasks.
Some methodologies, like CodeV-R1~\citep{zhu2025codevr1} and SpecLoop~\citep{chang2026specloop}, use \emph{round-trip reasoning} for dataset synthesis or specification refinement. We evaluate \emph{round-trip} as a task protocol using formal verification as an evaluation method. Specifically, formal verification uses mathematical methods to check whether generated RTL matches a reference design’s behavior. The \emph{round-trip protocol} together with the formal verification-based methodology significantly expands the benchmark scale.

\textbf{Project-level scope.} Pioneering works have made considerable efforts on project-level tasks and are mainly limited to benchmark scale. CVDP~\citep{pinckney2025cvdp} provides 166 agentic problems, with 37 specification-to-project and 26 module-reuse cases.
Spec2RTL-Agent~\citep{yu2025spec2rtl} explores specification-to-project generation based on three cases. 
RealBench~\citep{jin2025realbench} provides two project-level cases. HWE-Bench~\citep{cui2026hwebench} offers 417 repository issue-fixing cases based on 6 GitHub repositories. \systemname provides significantly more benchmark cases (1k+) from 50+ GitHub and OpenCores' projects. 

\begin{table}[t]
\vspace{-.3in}
  \centering
  \usetikzlibrary{patterns}
  \newlength{\blockedVerilogWidth}
  \newlength{\blockedRtllmWidth}
  \newlength{\blockedCvdpWidth}
  \newlength{\blockedRealbenchWidth}
  \newlength{\blockedLocationWidth}
  \newlength{\blockedHweWidth}
  \newcommand{\blockedcell}[1]{%
    \cellcolor{red!6}%
    \begin{tikzpicture}[baseline=0pt]
      \path[use as bounding box] (0,-0.4em) rectangle (#1,0.9em);
      \begin{pgfinterruptboundingbox}
        \path[pattern=north east lines,pattern color=gray!35]
          (-\tabcolsep,-0.4em) rectangle (#1+\tabcolsep,0.9em);
      \end{pgfinterruptboundingbox}
    \end{tikzpicture}%
  }

  \caption{Comparison with representative RTL benchmarks on task coverage and scale.}
  \label{tab:benchmark-comparison}
  \scriptsize
  \settowidth{\blockedVerilogWidth}{\textbf{VerilogEval}}
  \settowidth{\blockedRtllmWidth}{\textbf{RTLLM 2.0}}
  \settowidth{\blockedCvdpWidth}{\textbf{CVDP}}
  \settowidth{\blockedRealbenchWidth}{\textbf{RealBench}}
  \settowidth{\blockedLocationWidth}{\textbf{Location}}
  \settowidth{\blockedHweWidth}{\textbf{HWE-Bench}}
  \setlength{\tabcolsep}{3pt}
  \renewcommand{\arraystretch}{1.15}
  \resizebox{\linewidth}{!}{%
  \begin{tabular}{@{}lll*{7}{c}@{}}
    \toprule
    \multirow{-3}{*}{\textbf{Protocol}} & \multirow{-3}{*}{\textbf{Scope}} & \multirow{-3}{*}{\textbf{Type}} &
    \shortstack{\textbf{VerilogEval}\\\citeyearpar{liu2023verilogeval}} &
    \shortstack{\textbf{RTLLM 2.0}\\\citeyearpar{liu2025openllmrtl}} &
    \shortstack{\textbf{CVDP}\\\citeyearpar{pinckney2025cvdp}} &
    \shortstack{\textbf{RealBench}\\\citeyearpar{jin2025realbench}} &
    \shortstack{\textbf{Location}\\\textbf{is Key}\\\citeyearpar{yao2024location}} &
    \shortstack{\textbf{HWE-Bench}\\\citeyearpar{cui2026hwebench}} &
    \multirow{-3}{*}{\shortstack{\textbf{Ours}}} \\
    \midrule
    \multirow{6}{*}{Direct} & \multirow{3}{*}{Module-level} & Generation & \cellcolor{softgreen!5.6}156 & \cellcolor{softgreen!3.2}50 & \cellcolor{softgreen!4}78 & \cellcolor{softgreen!3.5}60 & \blockedcell{\blockedLocationWidth} & \blockedcell{\blockedHweWidth} & \cellcolor{softgreen!12}\textbf{705} \\
     &  & Completion & \blockedcell{\blockedVerilogWidth} & \blockedcell{\blockedRtllmWidth} & \cellcolor{softgreen!3.4}94 & \blockedcell{\blockedRealbenchWidth} & \blockedcell{\blockedLocationWidth} & \blockedcell{\blockedHweWidth} & \cellcolor{softgreen!12}\textbf{1{,}296} \\
     &  & Debugging & \blockedcell{\blockedVerilogWidth} & \blockedcell{\blockedRtllmWidth} & \cellcolor{softgreen!3}36 & \blockedcell{\blockedRealbenchWidth} & \cellcolor{softgreen!3.6}102 & \blockedcell{\blockedHweWidth} & \cellcolor{softgreen!12}\textbf{1{,}564} \\
    \cmidrule(lr){2-10}
     & \multirow{3}{*}{Project-level} & Generation & \blockedcell{\blockedVerilogWidth} & \blockedcell{\blockedRtllmWidth} & \cellcolor{softgreen!10.3}37 & \cellcolor{softgreen!3.4}4 & \blockedcell{\blockedLocationWidth} & \blockedcell{\blockedHweWidth} & \cellcolor{softgreen!12}\textbf{50} \\
     &  & Completion & \blockedcell{\blockedVerilogWidth} & \blockedcell{\blockedRtllmWidth} & \cellcolor{softgreen!5.8}26 & \blockedcell{\blockedRealbenchWidth} & \blockedcell{\blockedLocationWidth} & \blockedcell{\blockedHweWidth} & \cellcolor{softgreen!12}\textbf{110} \\
     &  & Debugging & \blockedcell{\blockedVerilogWidth} & \blockedcell{\blockedRtllmWidth} & \cellcolor{softgreen!3.8}11 & \blockedcell{\blockedRealbenchWidth} & \blockedcell{\blockedLocationWidth} & \cellcolor{softgreen!10.4}417 & \cellcolor{softgreen!12}\textbf{827} \\
    \midrule
    \multirow{6}{*}{Round-trip} & \multirow{3}{*}{Module-level} & Generation & \blockedcell{\blockedVerilogWidth} & \blockedcell{\blockedRtllmWidth} & \blockedcell{\blockedCvdpWidth} & \blockedcell{\blockedRealbenchWidth} & \blockedcell{\blockedLocationWidth} & \blockedcell{\blockedHweWidth} & \cellcolor{softgreen!12}\textbf{9{,}696} \\
     &  & Completion & \blockedcell{\blockedVerilogWidth} & \blockedcell{\blockedRtllmWidth} & \blockedcell{\blockedCvdpWidth} & \blockedcell{\blockedRealbenchWidth} & \blockedcell{\blockedLocationWidth} & \blockedcell{\blockedHweWidth} & \cellcolor{softgreen!12}\textbf{1{,}669} \\
     &  & Debugging & \blockedcell{\blockedVerilogWidth} & \blockedcell{\blockedRtllmWidth} & \blockedcell{\blockedCvdpWidth} & \blockedcell{\blockedRealbenchWidth} & \blockedcell{\blockedLocationWidth} & \blockedcell{\blockedHweWidth} & \cellcolor{softgreen!12}\textbf{1{,}648} \\
    \cmidrule(lr){2-10}
     & \multirow{3}{*}{Project-level} & Generation & \blockedcell{\blockedVerilogWidth} & \blockedcell{\blockedRtllmWidth} & \blockedcell{\blockedCvdpWidth} & \blockedcell{\blockedRealbenchWidth} & \blockedcell{\blockedLocationWidth} & \blockedcell{\blockedHweWidth} & \cellcolor{softgreen!12}\textbf{72} \\
     &  & Completion & \blockedcell{\blockedVerilogWidth} & \blockedcell{\blockedRtllmWidth} & \blockedcell{\blockedCvdpWidth} & \blockedcell{\blockedRealbenchWidth} & \blockedcell{\blockedLocationWidth} & \blockedcell{\blockedHweWidth} & \cellcolor{softgreen!12}\textbf{618} \\
     &  & Debugging & \blockedcell{\blockedVerilogWidth} & \blockedcell{\blockedRtllmWidth} & \blockedcell{\blockedCvdpWidth} & \blockedcell{\blockedRealbenchWidth} & \blockedcell{\blockedLocationWidth} & \blockedcell{\blockedHweWidth} & \cellcolor{softgreen!12}\textbf{596} \\
    \bottomrule
  \end{tabular}}
  \vspace{-.3in}
\end{table}

\begin{figureblock}%
\begin{minipage}{\linewidth}
\centering\setlength{\abovecaptionskip}{3pt}
\vspace{-.2in}
\includegraphics[width=\linewidth]{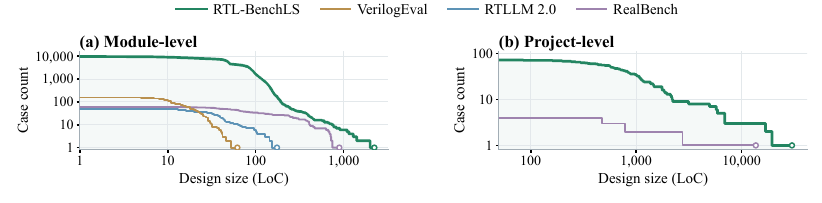}
\vspace{-.2in}
\caption{Comparison with representative benchmarks at one benchmark scale and design size.}
\label{fig:design-statistics}
\vspace{-.1in}
\end{minipage}%
\end{figureblock}

\section{RTL-BenchLS In Detail}
\label{sec:benchmark}

\subsection{Benchmark Tasks}
\label{sec:benchmark-tasks}

We define each task by its task type, task scope, and task protocol.

\textbf{Task type.} We define three task types to evaluate different capabilities of agents. \textbf{Type 1:} \emph{Generation} requires implementing RTL from a specification, including the required logic and interfaces. \textbf{Type 2:} \emph{Completion} requires recovering missing logic or modules from the available specification and retained RTL context, while preserving compatibility with the surrounding design. In existing work~\citep{liu2023verilogeval, lu2024rtllm, pinckney2025cvdp}, completion tasks are mainly defined as implementing the RTL logic based on the specification and an additional module header. In our work, we broaden the definition to fill-in-the-middle based on the surrounding context: LLMs may also be required to implement a missing code block or module in a project. \textbf{Type 3:} \emph{Debugging} requires correcting an implementation to satisfy its specification or resolve a reported issue. We collect debugging cases in two ways: \emph{(1) a mutant-insertion-based method for bug creation} and \emph{(2) real issue-fixing pairs} from realistic repository maintenance history.

\textbf{Task scope.} We separate two task scopes. \emph{(1) Module-level} tasks evaluate the LLM backbone's ability on detailed RTL logic in a one-shot generation manner. \emph{(2) Project-level} tasks target the agents' abilities in context-based reasoning and project development and maintenance, which requires iterative reasoning and tool calling. Agents may have to inspect the available project context, edit RTL, and invoke tools in isolated workspaces.

\textbf{Task protocol.} Existing benchmarks mainly rely on the \textbf{direct protocol}, which provides specifications, RTL context, or issue reports as input, and requires agents to generate desired output. In this protocol, input is predefined and fixed.  In this work, we introduce the \emph{round-trip} protocol. In the \textbf{round-trip protocol}, the agent generates an intermediate representation of the RTL design, which serves as input for the next reasoning step. For example, the agent may generate, complete, or debug RTL based on the intermediate representation and surrounding RTL context in a separate second step. We set two rules for the intermediate representation. \emph{(1) Leakage rules:} the agent can not leak hardware code, like Verilog or VHDL. We use a script to detect code leakage. \emph{(2) Code-based length limits:} the agent's intermediate representation should not exceed a length proportional to the RTL code. We use a \emph{length limit ratio} $r$ to control the length. This forces the agent to describe design functionality concisely while keeping the key semantics. This limitation also prevents the agent from describing overly detailed information. {This round-trip protocol is inherently harder because the agent has to perform two reasoning steps. (1) The agent should understand and compress the RTL code into an effective intermediate representation. (2) The agent has to recover from the compressed representations. The quality of the representation in step (1) also influences step (2).} Additionally, the round-trip protocol increases the number of usable designs for benchmarking without requiring an original specification--RTL pair.

We derive 12 tasks from these three dimensions in Table~\ref{tab:benchmark-comparison}. Direct Project-level debugging has three subtypes: mutant debugging, nonfunctional issue fixing, and functional issue fixing. Their scores are averaged in Table~\ref{tab:unified} and reported separately in Appendix~\ref{app:additional-results}. Task names follow \emph{task protocol + task scope + task type}. Appendix~\ref{app:tasks} lists the detailed task definitions. \emph{Mutant debugging} requires the agent to debug a mutant of the reference RTL based on the original specification. For real GitHub issue-fixing, we separate the fixings into \emph{nonfunctional issue fixing} and \emph{functional issue fixing}. {HWE-Bench~\citep{cui2026hwebench} mainly contributes functional issue-fixing tasks, where the commits change the function of the target modules. We find that some commits are functionally equivalent to the original code but still make meaningful modifications, such as removing redundant hardware logic and optimizing performance.} {We further include such \emph{nonfunctional} commits in our benchmark.} {To evaluate the non-functional issue-fixing task reasonably, we provide two evaluation metrics: (1) \emph{CommitMatch}-based verification, and (2) \emph{Checklist}-based judgment. Specifically, \emph{(1) CommitMatch} strictly matches the commit with the golden commit, which may reject alternative valid solutions. To tackle this limitation of method (1), we add \emph{(2) Checklist-based} judgment, which first extracts a list of checkpoints, which will be given to the strongest model, such as GPT-6.1 Sol, to judge whether the submission fulfills all requirements.}

\textbf{Evaluation process.} We utilize different evaluation metrics for tasks.
For \emph{module-level} generation and completion tasks, we report functional pass rate, following the metrics used in prior work~\citep {lu2024rtllm, liu2023verilogeval}. For \emph{project-level} tasks, we utilize testpoint-based evaluation metrics. Project-level tasks are significantly harder for agents to implement fully correctly. But the agents can still make meaningful progress. We use the official testbenches and formal verification on key submodules to construct meaningful testpoints to measure progress in project generation. The metric is thus the passed testpoints / total testpoints for a project. For \emph{design debugging}, we set \emph{easy} and \emph{hard} modes. \emph{Easy} mode provides the bug location, and the agent is only required to fix the bugs. For \emph{hard} mode, the agent has to identify the bug and fix it.

\subsection{Benchmark Data Source}
\label{sec:benchmark-data}

\textbf{Data composition.} We collect data from a broad range of sources for \emph{module-level} and \emph{project-level} tasks.
\emph{(1) On the module level.} \emph{For the Direct protocol,} we collect 700+ qualified designs. \emph{For the Round-trip protocol,} we collect 9k+ RTL designs, including 900+ from 25 open-source GitHub repositories and OpenCores IPs\footnote{OpenCores (\url{https://opencores.org/}) is a well-recognized community for reusable open-source IP. Its projects provide RTL sources and project metadata, such as design specifications and testbenches.} and 9000+ from four public RTL datasets~\citep{cui2024origen,auto-v-coder,allam2025rtlpp,zhang2024mg}. We apply strict filtering to the module-level RTL designs to remove duplicates and designs that fail compilation or formal-harness validation. The collection covers both combinational and sequential RTL, including logic, arithmetic, storage, finite-state control, interfaces, interconnects, processors, memory hierarchies, DSP and ML accelerators, security IP, and application-specific hardware.

\emph{(2) On the project level.} We collect 72 projects in total from high-quality OpenCores IP and highly starred GitHub repositories, including repository issues and fix histories. These projects cover complete processor, SoC, communication, memory, interconnect, cryptographic, video, and accelerator systems. \emph{For the direct tasks}, we curate 50 qualified projects that contain manually created project specifications.

\textbf{Benchmark organization.} Table~\ref{tab:benchmark-comparison} summarizes the case counts for each task. \emph{(1) On the module level.} We organize cases in two ways. \emph{Source-wise groups} organize RTL designs by source, indexed as S1--S5, to identify where the designs come from. Specifically, S1 contains repository designs, while S2--S5 correspond to the four public RTL datasets. We also divide cases into 20 slices to support evaluation in smaller batches. Each slice is balanced, with a matched proportion and design-size distribution to the full benchmark distribution. \emph{(2) On the project level.} We organize cases by their source project. Appendix~\ref{app:source-composition} provides the detailed list of project-level cases.

\section{RTL-BenchLS Generation Process}

\begin{figure}[!t]
    \vspace{-.2in}
  \centering
  \includegraphics[width=0.995\linewidth,trim=18bp 0bp 18bp 82bp,clip]{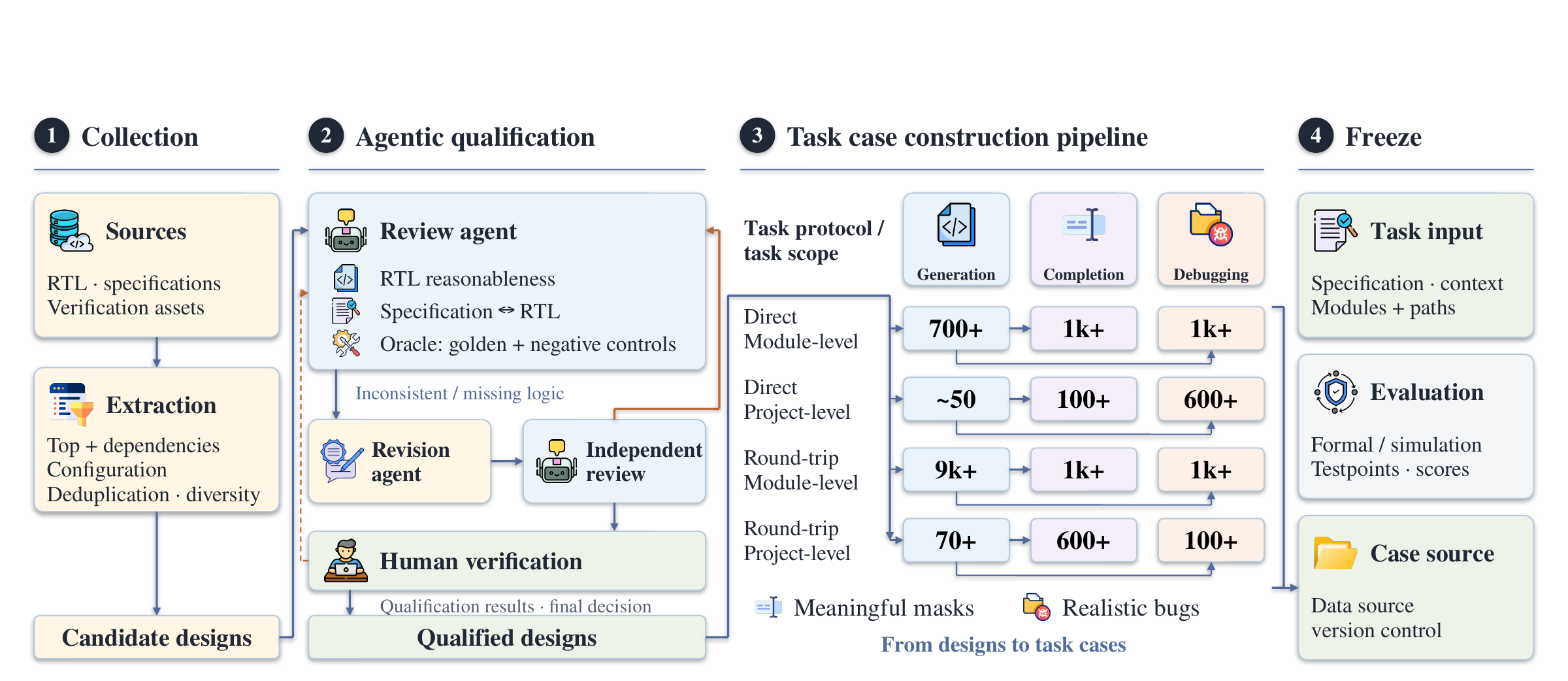}
  \vspace{-.15in}
  \caption{Data curation workflow in \systemname, with four stages from design collection to frozen task cases. \protect\filledcircled{1}~\textbf{Collection} collects RTL, available specifications, and verification assets from repositories and datasets. \protect\filledcircled{2}~\textbf{Agentic qualification} filters high quality designs and validate the specification quality. This step then constructs solid evaluation pipelines based on formal and simulation-based verification. \protect\filledcircled{3} then constructs task cases. All cases will be manually verified in step \protect\filledcircled{4}.}
  \label{fig:data-curation-workflow}
  \vspace{-.2in}
\end{figure}

Figure~\ref{fig:data-curation-workflow} summarizes four stages for collecting qualified RTL designs and constructing test cases. We use automated scripts and agentic frameworks to automate repeated collection and primary validation. We then manually verify the collected cases.

\filledcircled{1}~\textbf{Collect and extract designs.} We collect RTL, available specifications, and verification assets from repositories and datasets. During this process, we control source diversity and design size by removing duplicate designs and balancing lines of code (LoC) in the dataset. The filtering results are reported in Section~\ref{sec:appendix-filtering}. Two types of data cases are collected during this step: \emph{specification-RTL paired} cases and \emph{single RTL} cases. For \emph{specification-RTL paired} cases, we further conduct qualification verification, especially of the specification against the reference RTL code. The \emph{single RTL} designs are prepared for \emph{round-trip} tasks.

\filledcircled{2}~\textbf{Qualify designs and specifications.} In this stage, we first filter out the designs that are synthesizable and functionally verifiable. Specifically, designs must either pass the existing manually written testbench or be verifiable by formal verification tools. Then, we create the mutants from reference RTL code and validate that the testbench or formal scripts can successfully detect the mutant bugs. After the design qualification step, we further conduct agentic validation on specification quality. Motivated by RTL-BenchMT~\citep{wang2026rtl}, we design specification agents that review RTL behavior, check specifications against RTL, and diagnose inconsistent or missing requirements. Revisions undergo independent review, followed by final human verification. We only keep the specification-RTL pairs that pass the validation process. {Specifically, for module-level designs, the detailed behavior, like reset conditions and module interfaces, is clearly specified, so that each specification clearly refers to one RTL module. For \emph{project-level} designs, one specification may correspond to several valid RTL implementations, with different {configurations}. During \emph{project-level} task evaluation, the specification and one specific \emph{configuration} will be explicitly given to the agent so that there is only one specific reference RTL implementation for verification.}

\filledcircled{3}~\textbf{Construct task cases.} Based on the qualified cases, we first construct \textbf{type 1: design generation} tasks. The \emph{single RTL} designs are utilized to construct \emph{round-trip} tasks. The \emph{specification-RTL pairs} are utilized for \emph{direct} tasks. For \emph{design generation} tasks, the input is a design specification, and the required output is an RTL implementation. For \emph{module-level} generation tasks, only one verification against the targeted top module is conducted. For \emph{project-level} design implementation tasks, the evaluation involves multiple testbenches or formal verification against important sub-modules within the project. We define each verification step for the project as a ``\emph{testpoint}.'' The project-level design generation task is significantly challenging, but passing testpoints still effectively reveals the agent's progress on the project implementation. \emph{Type 2: design completion} and \emph{Type 3: design debugging} are mainly derived from \emph{Type 1: design generation} cases. Briefly, we construct \emph{design completion} tasks by masking an important block within the module or an important module within a project. We construct \emph{design debugging} tasks by creating mutants for the reference module or project. {Specifically, one RTL design contributes only one task case for \emph{Type 1: design generation}, and multiple cases for \emph{Type 2} and \emph{Type 3}}. Detailed task formulations are included in Appendix~\ref{app:data}.

\filledcircled{4}~\textbf{Admit and freeze cases.} This is a human-led stage, where we double-check the task formulation, evaluation pipeline, and specification quality. Only manually verified cases are accepted in our final benchmark dataset.

{\textbf{Data contamination analysis.}} Data contamination, or overfitting to public benchmarks, is a long-standing challenge~\cite{wang2026rtl}. Since our module-level generation tasks are derived from the existing open-source dataset, we also provide a data contamination analysis to ensure a low risk of contamination (Appendix~\ref{sec:data-contamination-analysis}).

\section{Experiments}
\label{sec:experiments}

\subsection{Experimental Setup}
\label{sec:setup}

\paragraph{Models and harness.} We evaluate both the commercial models and open-source models on our benchmark cases.
For commercial models, we select GPT-5.6 Sol, Claude Sonnet 4.5, and Claude Haiku 4.5. For open-source models, we select DeepSeek V3.2, Mistral Large, Qwen3.5-397B, and Llama 3.3 70B.
\emph{Project-level} tasks require an agentic harness. For commercial models, we use Codex CLI for GPT-5.6 and 5.4, and Claude Code for Claude Sonnet-4.5 and Haiku-4.5. For open-source models, we include Mistral Large, Qwen 3.5, DeepSeek-v3.2, and Llama-3.3. Specifically, we use the open-source harness framework OpenHands for the open-source models.

\textbf{Evaluation and verification.} The metrics for different tasks are introduced in the \textbf{evaluation process} of Section~\ref{sec:benchmark-tasks}. Our functional verification relies on formal verification and testbench-based simulation. For designs with manually written testbenches (from GitHub repositories or OpenCores projects), we reuse the qualified testbench for simulation. For designs without testbenches, we utilize the formal verification tools to verify the generated RTL against the reference RTL code. \emph{Each task is evaluated with repetition time $n=3$.}
For tasks at \emph{round-trip protocol}, we set \emph{length limit ratio} $r=1.0$ for \emph{module-level} tasks and $r=0.5$ for \emph{project-level} tasks. We set different ratios for the module and project task scopes because \emph{project-level} tasks usually have more available surrounding context. Appendix~\ref{app:additional-results} provides further exploration of round-trip module-level tasks. For project-level tasks, agents are given a running budget of \emph{32 agent iterations} under a \emph{30-minute} time limit for easy and medium cases and \emph{64 agent iterations} under a \emph{60-minute} time limit for difficult cases. We monitor runtime and confirm that these budgets are sufficient for agents to complete the project design tasks and submit for the final evaluation.

\textbf{Evaluation subsets on round-trip tasks.} Due to the large scale of the data cases and expensive API cost, for round-trip module-level tasks, we use slices~01, 05, 10, 15, and~20 for evaluation. In Appendix~\ref{app:slice-representativeness}, we show that each slice is balanced and has a similar data distribution. 

\subsection{Benchmarking Results}
\label{sec:main-results}

In this section, we mainly provide overall benchmarking results. Appendix~\ref{app:additional-results} provides more detailed results and analysis.
Table~\ref{tab:unified} reports functional scores for 12 combinations of task protocol, task scope, and task type. The three Direct Project-level debugging subtypes are averaged. \emph{Direct Module-level generation}, a long-standing standard task, shows the highest performance among the models. GPT-5.6 Sol demonstrates the strongest ability across tasks, with strong generalization across scopes and agentic capabilities for project-level hardware implementation. Generally, the \emph{project-level} and \emph{round-trip} tasks are harder for LLM agents.

\begin{table}[!ht]
\vspace{-.2in}
  \centering
  \caption{Main results organized by task protocol, task scope, and task type.}
  \label{tab:unified}
  \footnotesize
  \setlength{\tabcolsep}{2.1pt}
  \renewcommand{\arraystretch}{1.00}
  \resizebox{\textwidth}{!}{%
  \begin{tabular}{lllrrrrrrrr}
    \toprule
    \textbf{Task protocol} & \textbf{Task scope} & \textbf{Task type} &
    \shortstack{\textbf{GPT}\\\textbf{5.6 Sol}} &
    \shortstack{\textbf{GPT}\\\textbf{5.4}} &
    \shortstack{\textbf{Sonnet}\\\textbf{4.5}} &
    \shortstack{\textbf{Haiku}\\\textbf{4.5}} &
    \shortstack{\textbf{DeepSeek}\\\textbf{V3.2}} &
    \shortstack{\textbf{Mistral}\\\textbf{Large}} &
    \shortstack{\textbf{Qwen}\\\textbf{3.5}} &
    \shortstack{\textbf{Llama}\\\textbf{3.3}} \\
    \midrule
    \multirow{6}{*}{Direct} & \multirow{3}{*}{Module-level} & Generation & 74.2 & 54.1 & 54.6 & 38.6 & 46.4 & 26.1&  39.1 & 19.7 \\
    &  & Completion & {53.3} & 40.0 & \textbf{67.2} & {30.6} & 30.6 & 16.7 & 31.7 & 11.1 \\
     &  & Debugging & {63.2} & 52.3 & {60.0} & \textbf{70.0} & {56.7} & 2.3 & 16.7 & 13.8 \\
    \cmidrule(lr){2-11}
     & \multirow{3}{*}{Project-level} & Generation & 20.8 & 13.0 & 21.7 & 15.6 & {14.4} & 6.5 & 8.3 & 3.1 \\
     &  & Completion & 37.5 & 25.0 & 33.3 & {25.0} & 36.1 & 16.7 & 16.9 &  1.7 \\
     &  & Debugging & 34.5 & 34.3 & 37.6 & 32.5 & 28.2 & 23.0 & 16.7 & 10.0 \\
    \midrule
    \multirow{6}{*}{Round-trip} & \multirow{3}{*}{Module-level} & Generation & 44.0 & 35.0 & {39.9} & 35.7 & 33.0 & 24.5 & 33.2 & 26.0 \\
     &  & Completion & 29.1 & 22.4 & 21.5 & 22.8 & 22.7 & 18.0 & {22.9} & 23.4 \\
     &  & Debugging & 74.1 & 63.6 & {70.6} & 63.4 & 57.8 & 33.6 & 55.2 & 41.3 \\
    \cmidrule(lr){2-11}
    & \multirow{3}{*}{Project-level} & Generation &  27.5 & 6.2 & {10.0} & 2.4 & 1.3 & 1.6 & 3.7 & 2.3 \\
     &  & Completion & 26.2 & 15.0 & {18.8} & 11.4 & 18.9 & 4.3 & 11.2 & 1.7 \\
     &  & Debugging & 32.5 & 23.8 & {36.2} & 27.5 & 22.9 & 10.3 & 16.2 & 11.9 \\
    \bottomrule
  \end{tabular}
  }
  \vspace{2pt}
  \parbox{0.96\textwidth}{\raggedright\scriptsize}
    \vspace{-.2in}
\end{table}

\textbf{Task scope and task type.} On \emph{module level}, the \emph{design completion} task is generally harder than other task types. For example, Direct and Round-trip Module-level completion show lower performance across models. This suggests the agent is less capable of context understanding and reasoning than of implementing a hardware design directly from a given specification. On the \emph{project level}, the generation tasks are significantly harder than other task types. For Direct and Round-trip Project-level generation, we report testpoint pass rates rather than a ``{full pass} rate.'' The testpoint pass rates are more informative than the full pass rate because many models' full pass rates are close to 0. Even the testpoint pass rates are clearly lower than other task types across models. This demonstrates the current agents' limitation on project-level tasks.
Appendix~\ref{sec:token-consumption} further elaborates on the LLMs' limited ability on project-level tasks based on the token consumption statistics.
\textbf{Task protocol.} The \emph{Round-trip} protocol on \emph{generation} and \emph{completion} task types usually scores below their \emph{direct} counterparts. This suggests the agent is less capable of RTL design summarization than directly generating hardware RTL code. In the Appendix Table~\ref{app:representation-analysis}, we further analyze the influence of increasing ratio on performance for round-trip tasks.

\subsection{Benchmark Comparison}
\label{sec:benchmark-task-comparison}

In this section, we compare \emph{RTL-BenchLS} with existing module-level benchmarks along \emph{task protocol} and \emph{case difficulty}. Table~\ref{tab:challenging-comparison} shows performance comparisons across benchmark cases and protocols. Rows with the same benchmark report the performance under the same benchmark cases with different protocols (\emph{Direct} v.s. \emph{Round-trip}). We compare cases in \emph{Direct Module-level generation} of \systemname with RTLLM~\citep{lu2024rtllm} and VerilogEval~\citep{liu2023verilogeval}.

For the \textbf{task-protocol} comparison, \emph{Round-trip} pass rates are lower than \emph{Direct} pass rates among models within each benchmark. This phenomenon demonstrates that our proposed \emph{round-trip} tasks are more challenging than the standard \emph{direct} protocol. For the \textbf{case-difficulty} comparison, \emph{Direct} scores among models on \systemname cases are below the corresponding RTLLM and VerilogEval cases. This suggests our benchmark cases are more challenging under the same task protocol. Additionally, we include new evaluation results for downstream RTL-specific coders~\citep{zhu2025codevr1} fine-tuned on the RTL training dataset. The RTL-specific coders also perform worse on \systemname. For round-trip tasks, the RTL-specific coders' near-zero performance implies that these downstream models cannot generalize to round-trip tasks involving RTL reasoning and summarization.

\begin{table}[t]
    \vspace{-.2in}
  \centering
  \caption{Comparison with module-level benchmarks on cases and task design.}
  \label{tab:challenging-comparison}
  \scriptsize
  \setlength{\tabcolsep}{2.4pt}
  \resizebox{\textwidth}{!}{%
  \begin{tabular}{llrrrrrrrr}
    \toprule
    \textbf{Benchmark} & \textbf{Task protocol} & \textbf{GPT-5.4} & \textbf{Sonnet} & \textbf{Haiku} & \textbf{DeepSeek} & \textbf{Qwen} & \textbf{Llama} & \shortstack{\textbf{CodeV-R1}\\\textbf{(RL)}} & \shortstack{\textbf{QiMeng-}\\\textbf{SALV}} \\
    \midrule
    \multirow{2}{*}{RTLLM}
      & Direct & 74.2 & 68.2 & {74.0} & 62.3 & 56.7 & 49.8 & 68.0 & 62.0 \\
      
      & Round-trip & {54.9} & 36.5 & 42.7 & 29.7 & 11.2 & 13.4 & 0.1 & 0.0 \\
    \midrule
    \multirow{2}{*}{VerilogEval}
      & Direct & 82.1 & {80.1} & 79.5 & 70.5 & 58.3 & 53.2 & 68.8 & 57.1 \\
      & Round-trip & 64.1 & {62.2} & 50.0 & 58.3 & 30.1 & 12.2 & 0.2 & 0.1 \\
    \midrule
    \multirow{2}{*}{\textbf{RTL-BenchLS}}
      & Direct & 54.1 & 54.6 & 38.6 & 46.4 & 39.1 & 19.7 & 45.4 & 42.0 \\
      & Round-trip & 39.7 & {40.3} & 26.8 & 33.6 & 27.3 & 12.5 & 1.6 & 1.3 \\
    \bottomrule
  \end{tabular}}
    \vspace{-.2in}
\end{table}

\section{Discussion}

We analyze failure modes (Section~\ref{sec:overall-failure}) and performance across design sizes (Section~\ref{sec:design-scale}).
Appendices~\ref{sec:data-contamination-analysis} and~\ref{sec:round-trip-quality} examine data contamination and the quality of intermediate representations in round-trip reasoning, respectively.

\subsection{Failure Analysis}
\label{sec:overall-failure}

In this section, we further analyze failure reasons across tasks. 
Figure~\ref{fig:error-distribution} shows the proportion of failure reasons in each task. In the figure, a \emph{full pass} means a full pass on the task. There are some \emph{common failure reasons} shared across all tasks. Specifically, \emph{invalid/incomplete output} means the required RTL submission is incomplete or malformed (project-level tasks usually require the agent to submit in a required format). \emph{Compile/elaboration failed} means that the RTL code has syntax errors and is rejected by tools. \emph{Interface/integration error} refers to the port or module header definition mismatching the requirement. \emph{Verification failure} means the RTL code passes syntax checks but fails functional verification. There are also some \emph{task-specific} failures. For example, a \emph{partial pass} is specifically used in project-level tasks, where the generated RTL projects partially pass some testpoints. \emph{Abstraction rejected} only exists in a round-trip task, where the intermediate representation violates code-leakage or length rules. \emph{Localization failed} only exists in debugging tasks, where the agent wrongly identifies the fault location. The results suggest that verification failures are common in module-level completion and debugging, and compilation failures are prominent in direct project-level generation. \emph{Abstraction rejection} is substantial in \emph{Round-trip Project-level} tasks.

\begin{figureblock}%
\vspace{-.2in}
  \begin{minipage}{\textwidth}
  \centering
  \includegraphics[width=\textwidth]{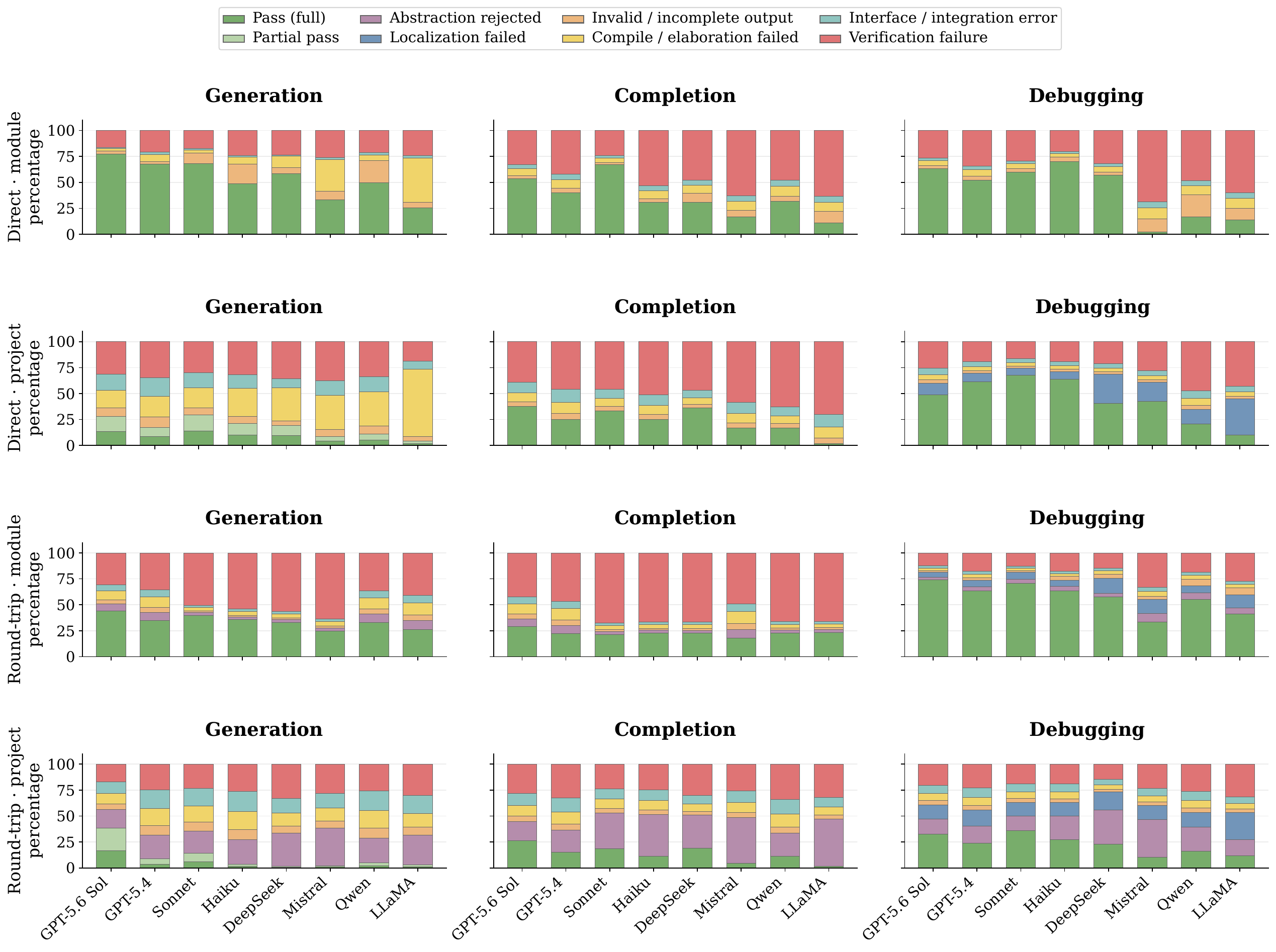}
  \vspace{-.2in}
  \caption{Error distributions on different tasks.}
  \label{fig:error-distribution}
  \end{minipage}
\end{figureblock}

\subsection{Benchmarking Results across Design Size}
\label{sec:design-scale}

Figure~\ref{fig:loc-overall} summarizes performance across design sizes for each task. \emph{Design generation} generally becomes harder as designs grow. Performance on \emph{project-level generation} tasks, including the \emph{direct} and \emph{round-trip} protocols, declines as LoC increases. For \emph{module-level} tasks, all models show relatively high performance on designs with LoC less than 100, and performance drops quickly as design size increases beyond 200. Compared with generation tasks, design size has less influence on completion and debugging tasks.

\begin{figureblock}%
  \begin{minipage}{\textwidth}
    \centering
    \includegraphics[width=\linewidth]{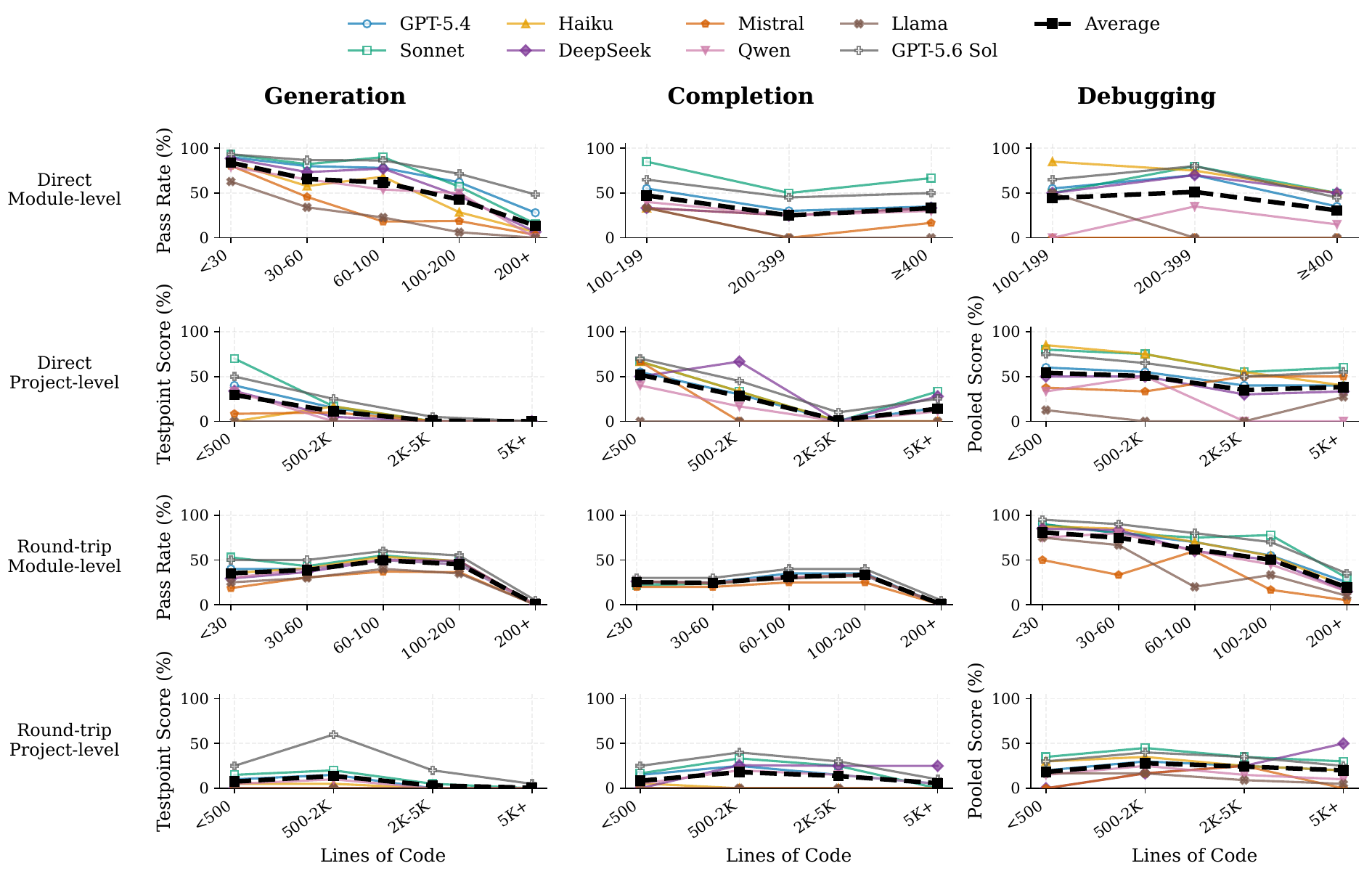}
    \caption{Performance across design size on different tasks. Curves show model results by LoC bin.}
    \label{fig:loc-overall}
  \end{minipage}
\end{figureblock}

\section{Conclusion}
\label{sec:conclusion}

We presented \systemname, a large-scale RTL benchmark with 17K+ cases and 12 tasks, spanning generation, completion, and debugging task types at the Module and project levels under Direct and Round-trip task protocols. Experiments show that our module-level cases are sufficiently challenging, and project-level tasks expose obvious limitations of current agentic frameworks. The \emph{round-trip} tasks reveal a significant gap between models’ ability to represent hardware behavior and their ability to generate RTL.
This new benchmark, at unprecedented scale, will help evaluate academic work and industrial solutions and improve LLM/agent solutions for hardware RTL design.
\section*{{Impact Statement}}

This benchmark paper collects all RTL designs and evaluation assets from open-source platforms, including GitHub repositories and OpenCores. The intended release contains public sources and
evaluation assets. The authors will verify provenance and license terms before publication.

\section*{{AI Use Disclosure}}

We use Generative AI tools, including ChatGPT and Codex, to help with experiment design, code implementation, literature discovery, figure plotting, and manuscript editing. The human authors remain responsible for reviewing and validating the final references, figures, code, and
benchmark artifacts. We have manually drafted the final manuscript.

\bibliographystyle{iclr2027_conference}
\bibliography{ref}

@article{wang2026rtl,
  title={RTL-BenchMT: Dynamic Maintenance of RTL Generation Benchmark Through Agent-Assisted Analysis and Revision},
  author={Wang, Jing and Liu, Shang and Zhou, Hangan and Xie, Zhiyao},
  journal={arXiv preprint arXiv:2605.15537},
  year={2026}
}

@article{pei2024betterv,
  title={Betterv: Controlled verilog generation with discriminative guidance},
  author={Pei, Zehua and Zhen, Hui-Ling and Yuan, Mingxuan and Huang, Yu and Yu, Bei},
  journal={arXiv preprint arXiv:2402.03375},
  year={2024}
}

@article{yao2024location,
  title={Location is Key: Leveraging Large Language Model for Functional Bug Localization in Verilog},
  author={Yao, Bingkun and Wang, Ning and Zhou, Jie and Wang, Xi and Gao, Hong and Jiang, Zhe and Guan, Nan},
  journal={arXiv preprint arXiv:2409.15186},
  year={2024}
}

@article{ho2024verilogcoder,
  title={VerilogCoder: Autonomous Verilog Coding Agents with Graph-based Planning and Abstract Syntax Tree (AST)-based Waveform Tracing Tool},
  author={Ho, Chia-Tung and Ren, Haoxing and Khailany, Brucek},
  journal={arXiv preprint arXiv:2408.08927},
  year={2024}
}

@article{cui2024origen,
  title={OriGen: Enhancing RTL Code Generation with Code-to-Code Augmentation and Self-Reflection},
  author={Cui, Fan and Yin, Chenyang and Zhou, Kexing and Xiao, Youwei and Sun, Guangyu and Xu, Qiang and Guo, Qipeng and Song, Demin and Lin, Dahua and Zhang, Xingcheng and others},
  journal={arXiv preprint arXiv:2407.16237},
  year={2024}
}

@article{thakur2024verigen,
  title={Verigen: A large language model for verilog code generation},
  author={Thakur, Shailja and Ahmad, Baleegh and Pearce, Hammond and Tan, Benjamin and Dolan-Gavitt, Brendan and Karri, Ramesh and Garg, Siddharth},
  journal={ACM Transactions on Design Automation of Electronic Systems},
  volume={29},
  number={3},
  pages={1--31},
  year={2024},
  publisher={ACM New York, NY}
}

@article{pinckney2024revisiting,
  title={Revisiting VerilogEval: Newer LLMs, In-Context Learning, and Specification-to-RTL Tasks},
  author={Pinckney, Nathaniel and Batten, Christopher and Liu, Mingjie and Ren, Haoxing and Khailany, Brucek},
  journal={arXiv preprint arXiv:2408.11053},
  year={2024}
}

@inproceedings{liu2024rtlcoder,
  title={Rtlcoder: Outperforming gpt-3.5 in design rtl generation with our open-source dataset and lightweight solution},
  author={Liu, Shang and Fang, Wenji and Lu, Yao and Zhang, Qijun and Zhang, Hongce and Xie, Zhiyao},
  booktitle={2024 IEEE LLM Aided Design Workshop (LAD)},
  pages={1--5},
  year={2024},
  organization={IEEE}
}

@inproceedings{liu2023verilogeval,
  title={Verilogeval: Evaluating large language models for verilog code generation},
  author={Liu, Mingjie and Pinckney, Nathaniel and Khailany, Brucek and Ren, Haoxing},
  booktitle={2023 IEEE/ACM International Conference on Computer Aided Design (ICCAD)},
  pages={1--8},
  year={2023},
  organization={IEEE}
}

@inproceedings{lu2024rtllm,
  title={Rtllm: An open-source benchmark for design rtl generation with large language model},
  author={Lu, Yao and Liu, Shang and Zhang, Qijun and Xie, Zhiyao},
  booktitle={2024 29th Asia and South Pacific Design Automation Conference (ASP-DAC)},
  pages={722--727},
  year={2024},
  organization={IEEE}
}

@inproceedings{pearce2020dave,
  title={Dave: Deriving automatically verilog from english},
  author={Pearce, Hammond and Tan, Benjamin and Karri, Ramesh},
  booktitle={ Workshop on Machine Learning for CAD (MLCAD)},
  year={2020}
}

@article{chang2023chipgpt,
 author = {Chang, Kaiyan and Wang, Ying and Ren, Haimeng and Wang, Mengdi and Liang, Shengwen and Han, Yinhe and Li, Huawei and Li, Xiaowei},
 journal={ arXiv preprint arXiv:2305.14019 },
 title = {ChipGPT: How far are we from natural language hardware design},
 year = {2023}
}

@article{thakur2023autochip,
 author = {Thakur, Shailja and Blocklove, Jason and Pearce, Hammond and Tan, Benjamin and Garg, Siddharth and Karri, Ramesh},
 journal={ arXiv preprint arXiv:2311.04887 },
 title = {AutoChip: Automating HDL Generation Using LLM Feedback},
 year = {2023}
}

@inproceedings{auto-v-coder,
 author = {Gao, Mingzhe and Zhao, Jieru and Lin, Zhe and Ding, Wenchao and Hou, Xiaofeng and Feng, Yu and Li, Chao and Guo, Minyi},
 booktitle={ International Conference on Computer Design (ICCD) },
 title = {AutoVCoder: A Systematic Framework for Automated Verilog Code Generation using LLMs},
 year = {2024}
}

@article{zhao2024mage,
  title={MAGE: A Multi-Agent Engine for Automated RTL Code Generation},
  author={Zhao, Yujie and Zhang, Hejia and Huang, Hanxian and Yu, Zhongming and Zhao, Jishen},
  journal={arXiv preprint arXiv:2412.07822},
  year={2024}
}

@article{kang2024fveval,
  title={FVEval: Understanding Language Model Capabilities in Formal Verification of Digital Hardware},
  author={Kang, Minwoo and Liu, Mingjie and Hamad, Ghaith Bany and Suhaib, Syed and Ren, Haoxing},
  journal={arXiv preprint arXiv:2410.23299},
  year={2024}
}

@inproceedings{zhang2024mg,
  title={Mg-verilog: Multi-grained dataset towards enhanced llm-assisted verilog generation},
  author={Zhang, Yongan and Yu, Zhongzhi and Fu, Yonggan and Wan, Cheng and Lin, Yingyan Celine},
  booktitle={2024 IEEE LLM Aided Design Workshop (LAD)},
  pages={1--5},
  year={2024},
  organization={IEEE}
}

@article{zhu2025codevr1,
  title={{QiMeng-CodeV-R1}: Reasoning-Enhanced Verilog Generation},
  author={Zhu, Yaoyu and Huang, Di and Lyu, Hanqi and Zhang, Xiaoyun and Li, Chongxiao and Shi, Wenxuan and Wu, Yutong and Mu, Jianan and Wang, Jinghua and Zhao, Yang and others},
  journal={arXiv preprint arXiv:2505.24183},
  year={2025}
}

@article{chang2026specloop,
  title={{SpecLoop}: An Agentic RTL-to-Specification Framework with Formal Verification Feedback Loop},
  author={Chang, Fu-Chieh and Yang, Yu-Hsin and Huang, Hung-Ming and Hsu, Yun-Chia and Lin, Yin-Yu and Tsai, Ming-Fang and Yang, Chun-Chih and Wu, Pei-Yuan},
  journal={arXiv preprint arXiv:2603.02895},
  year={2026}
}

@article{huang2025rtlspec,
  title={Assessing Large Language Models in Generating {RTL} Design Specifications},
  author={Huang, Hung-Ming and Yang, Yu-Hsin and Chang, Fu-Chieh and Hsu, Yun-Chia and Lin, Yin-Yu and Tsai, Ming-Fang and Yang, Chun-Chih and Wu, Pei-Yuan},
  journal={arXiv preprint arXiv:2512.00045},
  year={2025}
}

@article{pinckney2025cvdp,
  title={Comprehensive Verilog Design Problems: A Next-Generation Benchmark Dataset for Evaluating Large Language Models and Agents on {RTL} Design and Verification},
  author={Pinckney, Nathaniel and Deng, Christopher and Ho, Chia-Tung and Tsai, Yi-Dian and Liu, Mingjie and Zhou, Wenfei and Khailany, Brucek and Ren, Haoxing},
  journal={arXiv preprint arXiv:2506.14074},
  year={2025}
}

@article{liu2025deeprtl,
  title={{DeepRTL}: Bridging Verilog Understanding and Generation with a Unified Representation Model},
  author={Liu, Yuxiao and Xu, Chenxing and Zhou, Yanrui and Li, Zhen and Xu, Qiang},
  journal={arXiv preprint arXiv:2502.15832},
  year={2025}
}

@inproceedings{allam2025rtlpp,
  title={RTLPP: A Parallel Processing RTL Code Generation Framework Using LLMs},
  author={Allam, Ahmed and others},
  booktitle={Workshop on LLM-Aided Design (LAD)},
  year={2025}
}

@inproceedings{allam2024rtlrepo,
  title={RTL-Repo: A Benchmark for Evaluating LLMs on Large-Scale RTL Design Projects},
  author={Allam, Ahmed and others},
  booktitle={Workshop on LLM-Aided Design (LAD)},
  year={2024}
}

@misc{liu2025openllmrtl,
  title     = {{OpenLLM-RTL}: Open Dataset and Benchmark for {LLM}-Aided Design {RTL} Generation},
  author    = {Liu, Shang and Lu, Yao and Fang, Wenji and Li, Mengming and Xie, Zhiyao},
  year      = {2025},
  note      = {arXiv:2503.15112}
}

@misc{cui2026hwebench,
  title     = {{HWE-Bench}: Benchmarking {LLM} Agents on Real-World Hardware Bug Repair Tasks},
  author    = {Cui, Fan and Hou, Hongyuan and Luo, Zizhang and Yin, Chenyun and Liang, Yun},
  year      = {2026},
  note      = {arXiv:2604.14709}
}

@inproceedings{yosys,
  author    = {Wolf, Clifford and Glaser, Johann},
  title     = {{Yosys}: A Free {Verilog} Synthesis Suite},
  booktitle = {Proceedings of Austrochip 2013},
  pages     = {47--52},
  year      = {2013}
}

@article{jin2025realbench,
  title={{RealBench: Benchmarking Verilog Generation Models with Real-World IP Designs}},
  author={Jin, Pengwei and Huang, Di and Li, Chongxiao and Cheng, Shuyao and Zhao, Yang and Zheng, Xinyao and Zhu, Jiaguo and Xing, Shuyi and Dou, Bohan and Zhang, Rui and Du, Zidong and Guo, Qi and Hu, Xing},
  journal={arXiv preprint arXiv:2507.16200},
  year={2025}
}

@article{yu2025spec2rtl,
  title={{Spec2RTL-Agent: Automated Hardware Code Generation from Complex Specifications Using LLM Agent Systems}},
  author={Yu, Zhongzhi and Liu, Mingjie and Zimmer, Michael and Lin, Yingyan Celine and Liu, Yong and Ren, Haoxing},
  journal={arXiv preprint arXiv:2506.13905},
  year={2025}
}

\clearpage
\appendix
\section{Benchmark Construction Details}
\label{app:data}

\subsection{Verification Methods}
\label{app:verification-methods}

\paragraph{Testbench-based evaluation.}
We reuse upstream testbenches from the selected GitHub and OpenCores repositories, with support for Synopsys VCS and open-source Icarus Verilog. This flow mainly evaluates Project-level generation and also supports completion and debugging. We validate each testbench on the reference RTL. In project-level tasks, each testbench will be utilized as one testpoint.

\paragraph{Formal verification.}
Module-level tasks and formally checked Project-level completion and debugging tasks frequently use formal verification tools to check the equivalence of submitted RTL against the clean or developer-fixed reference. We support Cadence Conformal and Synopsys Formality for logical equivalence checking (LEC), Cadence JasperGold for sequential equivalence checking (SEC), and open-source Yosys~\citep{yosys}.

\subsection{Source Composition}
\label{app:source-composition}
\begin{wrapfigure}{R}{0.47\textwidth}
\vspace{-8pt}
\centering\setlength{\abovecaptionskip}{3pt}
\input{section/figures/design-types}
\end{wrapfigure}

Figure~\ref{fig:design-types} provides a brief status on the design types across our benchmark.
At \emph{module level}, we have five sources, denoted from S1 to S5. Specifically,
S1 contains 900+ designs from 25 GitHub repositories and spans processors, accelerators, networking blocks, and reusable hardware libraries. S2--S5 contribute 9k+ designs from OriGen, AutoVCoder, RTLPP, and MG-Verilog.
Table~\ref{tab:earlier-source-composition} shows the data composition on the module-level tasks. From the 10k+ cases, we further derive 700+ high-quality \emph{direct module-level generation} task cases.

\WFclear
\vspace{1.5\baselineskip}
\begin{tableblock}
  \begin{minipage}{\textwidth}
    \centering
    \caption{Data distribution on round-trip module-level tasks.}
    \label{tab:earlier-source-composition}
    \small
    \setlength{\tabcolsep}{5pt}
    \begin{tabular}{@{}llrrrl@{}}
      \toprule
      \textbf{Group} & \textbf{Source} & \textbf{Count} & \textbf{Avg. LoC} & \textbf{Max. LoC} & \textbf{Description} \\
      \midrule
      S1 & nvdla/hw & 404 & 41 & 479 & NVIDIA DLA \\
         & basic & 230 & 19 & 198 & General Verilog library \\
         & oh & 85 & 36 & 312 & Open hardware library \\
         & hdl & 44 & 28 & 165 & ADI HDL reference \\
         & openc910 & 43 & 86 & 421 & XuanTie C910 CPU \\
         & zet & 27 & 52 & 247 & x86 processor \\
         & v-ethernet & 17 & 68 & 285 & Ethernet MAC/PHY \\
         & e200 & 16 & 44 & 189 & HummingBird MCU \\
         & Other repositories (17) & 105 & 38 & 395 & CPUs, FPU, FIFO, etc. \\
      \cmidrule(lr){2-6}
         & \textbf{S1 subtotal} & \textbf{971} & \textbf{38} & \textbf{479} & \\
      \midrule
      S2 & OriGen & 2{,}708 & 72 & 189 & Augmented RTL corpus \\
      S3 & AutoVCoder & 2{,}468 & 78 & 331 & Auto Verilog generation \\
      S4 & RTLPP & 2{,}164 & 82 & 263 & Parallel-processing RTL \\
      S5 & MG-Verilog & 1{,}717 & 29 & 156 & Multi-grained dataset \\
      \midrule
         & \textbf{Total} & \textbf{10{,}028} & \textbf{63} & \textbf{479} & \\
      \bottomrule
    \end{tabular}
  \end{minipage}
\end{tableblock}

At \emph{project level}, we collect 70+ cases from 50+ starred GitHub repositories and OpenCore projects. Table~\ref{tab:project-case-inventory} lists the collected projects.

\begingroup
\footnotesize
\setlength{\tabcolsep}{4pt}
\renewcommand{\arraystretch}{1.08}
\begin{longtable}{@{}>{\raggedright\arraybackslash}p{6.6cm}>{\raggedright\arraybackslash}p{6.0cm}@{}}
\caption{List of collected projects for project-level generation.}\label{tab:project-case-inventory}\\
\toprule
\textbf{Case ID} & \textbf{Source project} \\
\midrule
\endfirsthead
\multicolumn{2}{c}{\textit{Table~\thetable{} continued}} \\
\toprule
\textbf{Case ID} & \textbf{Source project} \\
\midrule
\endhead
\midrule
\multicolumn{2}{r}{\textit{Continued on next page}} \\
\endfoot
\bottomrule
\endlastfoot
\nolinkurl{t1.4.upstream172.tiny_gpu} & \nolinkurl{github/adam-maj/tiny-gpu} \\
\nolinkurl{github/alexforencich/verilog-axi/axi_dma} & \nolinkurl{github/alexforencich/verilog-axi} \\
\nolinkurl{github/alexforencich/verilog-axi/axi_fifo} & \nolinkurl{github/alexforencich/verilog-axi} \\
\nolinkurl{github/alexforencich/verilog-axi/axi_register} & \nolinkurl{github/alexforencich/verilog-axi} \\
\nolinkurl{github/alexforencich/verilog-axi/axil_adapter} & \nolinkurl{github/alexforencich/verilog-axi} \\
\nolinkurl{github/alexforencich/verilog-axi/axil_cdc} & \nolinkurl{github/alexforencich/verilog-axi} \\
\nolinkurl{github/alexforencich/verilog-axi/axil_reg_if} & \nolinkurl{github/alexforencich/verilog-axi} \\
\nolinkurl{github/alexforencich/verilog-axi/axil_register} & \nolinkurl{github/alexforencich/verilog-axi} \\
\nolinkurl{github/alexforencich/verilog-ethernet/arp} & \nolinkurl{github/alexforencich/verilog-ethernet} \\
\nolinkurl{github/alexforencich/verilog-i2c/i2c_master_axil} & \nolinkurl{github/alexforencich/verilog-i2c} \\
\nolinkurl{github/alexforencich/verilog-i2c/i2c_master_wbs_16} & \nolinkurl{github/alexforencich/verilog-i2c} \\
\nolinkurl{github/alexforencich/verilog-i2c/i2c_master_wbs_8} & \nolinkurl{github/alexforencich/verilog-i2c} \\
\nolinkurl{github/alexforencich/verilog-i2c/i2c_slave_axil_master} & \nolinkurl{github/alexforencich/verilog-i2c} \\
\nolinkurl{github/alexforencich/verilog-i2c/i2c_slave_wbm} & \nolinkurl{github/alexforencich/verilog-i2c} \\
\nolinkurl{github/alexforencich/verilog-pcie/dma_if_axi} & \nolinkurl{github/alexforencich/verilog-pcie} \\
\nolinkurl{t2.3.github_alexforencich_verilog-uart} & \nolinkurl{github/alexforencich/verilog-uart} \\
\nolinkurl{github/analogdevicesinc/hdl/ad_mux} & \nolinkurl{github/analogdevicesinc/hdl} \\
\nolinkurl{github/analogdevicesinc/hdl/axi_dmac_transfer} & \nolinkurl{github/analogdevicesinc/hdl} \\
\nolinkurl{github/chipsalliance/cores-veer-eh1} & \nolinkurl{github/chipsalliance/cores-veer-eh1} \\
\nolinkurl{t1.4.upstream172.veer_el2} & \nolinkurl{github/chipsalliance/cores-veer-el2} \\
\nolinkurl{github/corundum/corundum/mqnic_core_axi} & \nolinkurl{github/corundum/corundum} \\
\nolinkurl{t2.3.github_dpretet_async_fifo} & \nolinkurl{github/dpretet/async_fifo} \\
\nolinkurl{t1.4.upstream172.axi_crossbar} & \nolinkurl{github/dpretet/axi-crossbar} \\
\nolinkurl{github/evensgn/risc-v-cpu} & \nolinkurl{github/evensgn/risc-v-cpu} \\
\nolinkurl{github/lawrie/fpga_pio} & \nolinkurl{github/lawrie/fpga_pio} \\
\nolinkurl{t1.4.upstream172.step_mips_lab10} & \nolinkurl{github/lvyufeng/step_into_mips} \\
\nolinkurl{t1.4.upstream172.step_mips_lab4} & \nolinkurl{github/lvyufeng/step_into_mips} \\
\nolinkurl{t1.4.upstream172.step_mips_lab5} & \nolinkurl{github/lvyufeng/step_into_mips} \\
\nolinkurl{t1.4.upstream172.step_mips_lab6} & \nolinkurl{github/lvyufeng/step_into_mips} \\
\nolinkurl{t1.4.upstream172.step_mips_lab7} & \nolinkurl{github/lvyufeng/step_into_mips} \\
\nolinkurl{t1.4.upstream172.step_mips_lab8} & \nolinkurl{github/lvyufeng/step_into_mips} \\
\nolinkurl{t1.4.upstream172.step_mips_lab9} & \nolinkurl{github/lvyufeng/step_into_mips} \\
\nolinkurl{t1.4.upstream172.schoolmips} & \nolinkurl{github/mipsfpga/schoolmips} \\
\nolinkurl{github/nonarkitten/amiga_replacement_project/cia8520} & \nolinkurl{github/nonarkitten/amiga_replacement_project} \\
\nolinkurl{github/pconst/basic_verilog/cookbook_des_crypt} & \nolinkurl{github/pconst/basic_verilog} \\
\nolinkurl{t2.3.github_secworks_aes} & \nolinkurl{github/secworks/aes} \\
\nolinkurl{t2.3.github_secworks_sha256} & \nolinkurl{github/secworks/sha256} \\
\nolinkurl{github/t-k-233/risc-v-single-cycle-cpu/riscv_core} & \nolinkurl{github/t-k-233/risc-v-single-cycle-cpu} \\
\nolinkurl{github/ultraembedded/riscv/riscv_tcm_top} & \nolinkurl{github/ultraembedded/riscv} \\
\nolinkurl{github/ultraembedded/riscv/riscv_top} & \nolinkurl{github/ultraembedded/riscv} \\
\nolinkurl{github/wangxuan95/fpga-ftdi245fifo/ftdi245fifo} & \nolinkurl{github/wangxuan95/fpga-ftdi245fifo} \\
\nolinkurl{t1.4.upstream172.uart2axi4} & \nolinkurl{github/wangxuan95/fpga-uart} \\
\nolinkurl{github/zipcpu/wbuart32/wbuart} & \nolinkurl{github/zipcpu/wbuart32} \\
\nolinkurl{opencores/aes_core/aes_cipher_top} & \nolinkurl{opencores/aes_core} \\
\nolinkurl{opencores/aes_core/aes_inv_cipher_top} & \nolinkurl{opencores/aes_core} \\
\nolinkurl{opencores/arithmetic_core_configurable_cordic_core_in_verilog} & \nolinkurl{opencores/arithmetic_core_configurable_cordic_core_in_verilog} \\
\nolinkurl{t1.4.upstream172.crcahb} & \nolinkurl{opencores/arithmetic_core_crcahb} \\
\nolinkurl{opencores/arithmetic_core_double_fpu_verilog} & \nolinkurl{opencores/arithmetic_core_double_fpu_verilog} \\
\nolinkurl{opencores/arithmetic_core_fpu/unique_tuples_16512_v1} & \nolinkurl{opencores/arithmetic_core_fpu} \\
\nolinkurl{t2.3.opencores_trig64_v1} & \nolinkurl{opencores/arithmetic_core_trigonometric_functions_degrees_in_double_fpu} \\
\nolinkurl{t2.3.opencores_gpio31_v1} & \nolinkurl{opencores/communication_controller_general-purpose_i-o_gpio_core} \\
\nolinkurl{t2.3.opencores_i2c_master_v1} & \nolinkurl{opencores/communication_controller_i2c_controller_core} \\
\nolinkurl{opencores/communication_controller_i2c_to_wb} & \nolinkurl{opencores/communication_controller_i2c_to_wb} \\
\nolinkurl{t2.3.opencores_simple_spi_v1} & \nolinkurl{opencores/communication_controller_spi_core} \\
\nolinkurl{t2.3.opencores_uart_to_bus_v1} & \nolinkurl{opencores/communication_controller_uart_to_bus} \\
\nolinkurl{t2.3.opencores_uartspi_8n1_v1} & \nolinkurl{opencores/communication_controller_uart_to_spi} \\
\nolinkurl{t2.3.opencores_lpc_host_v1} & \nolinkurl{opencores/communication_controller_wishbone_lpc_host_and_peripheral_bridge} \\
\nolinkurl{t2.3.opencores_aes128_pipeline_v1} & \nolinkurl{opencores/crypto_core_aes-128_encryption} \\
\nolinkurl{t1.4.upstream172.fast_aes_10cycle} & \nolinkurl{opencores/crypto_core_fast_aes-128_encryption_only_cores} \\
\nolinkurl{t1.4.upstream172.fast_aes_1cycle} & \nolinkurl{opencores/crypto_core_fast_aes-128_encryption_only_cores} \\
\nolinkurl{t1.4.upstream172.fast_aes_5cycle} & \nolinkurl{opencores/crypto_core_fast_aes-128_encryption_only_cores} \\
\nolinkurl{opencores/crypto_core_high_throughput_and_low_area_aes_core} & \nolinkurl{opencores/crypto_core_high_throughput_and_low_area_aes_core} \\
\nolinkurl{opencores/crypto_core_montgomery_modular_multiplier_and_exponentiator/modular_multiplier} & \nolinkurl{opencores/crypto_core_montgomery_modular_multiplier_and_exponentiator} \\
\nolinkurl{t2.3.opencores_present80_v1} & \nolinkurl{opencores/crypto_core_present_cipher_encryption_core} \\
\nolinkurl{t2.3.opencores_keccak512_high_v1} & \nolinkurl{opencores/crypto_core_sha3_keccak} \\
\nolinkurl{t2.3.opencores_sha256_v1} & \nolinkurl{opencores/crypto_core_sha_cores} \\
\nolinkurl{t2.3.opencores_pid32_v1} & \nolinkurl{opencores/dsp_core_pid_controller} \\
\nolinkurl{t2.3.opencores_sdram32_v1} & \nolinkurl{opencores/memory_core_8-16-32_bit_sdram_controller} \\
\nolinkurl{opencores/other_programmable_interval_timer} & \nolinkurl{opencores/other_programmable_interval_timer} \\
\nolinkurl{opencores/processor_turbo_8051/reference_8051} & \nolinkurl{opencores/processor_turbo_8051} \\
\nolinkurl{t2.3.opencores_memory_acs32_v1} & \nolinkurl{opencores/system_controller_memory_controller_ip_core} \\
\nolinkurl{t2.3.opencores_vga_rgb_v1} & \nolinkurl{opencores/video_controller_vga-lcd_controller} \\
\end{longtable}
\endgroup

\subsection{Module-level Design Filtering}\label{sec:appendix-filtering}

\begin{figure}[htbp]
  \centering
  \includegraphics[width=0.55\linewidth]{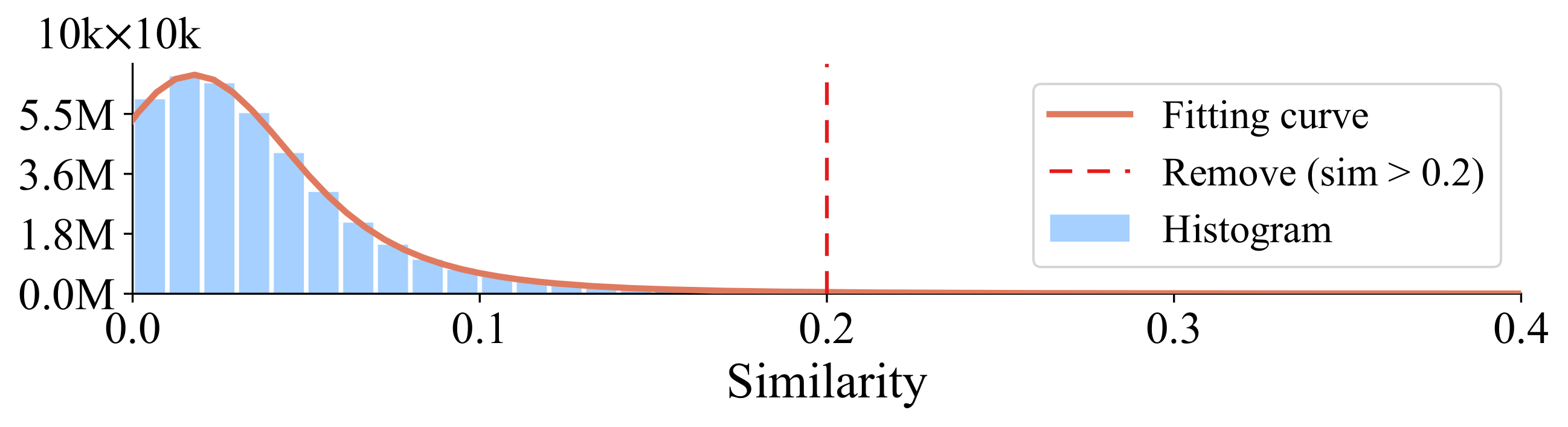}
  \caption{Pairwise similarity of collected Module-level RTL designs.}
  \label{fig:similarity-hist}
\end{figure}

We extract self-contained modules and their dependencies, normalize source packaging, and screen syntax, source size, and module hierarchy. We remove duplicate RTL, and near-duplicate screening uses cosine similarity between TF--IDF vectors of character 4-grams. Figure~\ref{fig:similarity-hist} summarizes the pairwise similarity distribution.

We review RTL behavior and specification consistency, including interfaces, widths, reset behavior, and corner cases. Flawed specifications are revised and independently reviewed. Unresolved specifications and designs are removed. Formal qualification checks whether the designs are compilable and functionally verifiable. Specifically, we formally verify the designs against functionally different mutants to validate that our formal verification pipeline is functioning correctly.

\clearpage
\subsection{Examples of Specifications}\label{app:specification-examples}

Figures~\ref{fig:module-specification-example} and
\ref{fig:project-specification-example} show examples for \emph{module-level} and \emph{project-level} specifications. During our case collection and validation process, we will strictly check the specification quality to eliminate ambiguities, such as input and output interfaces, clock and reset conditions, and other related configurations.

\begingroup
\usetikzlibrary{shadows}
\definecolor{specblue}{RGB}{71,103,145}
\definecolor{specteal}{RGB}{51,124,124}
\definecolor{specgray}{RGB}{105,112,121}

\begin{figureblock}%
\begin{minipage}{\textwidth}
\centering
\begin{tikzpicture}
\node[draw=specblue!65,fill=specblue!4,line width=.55pt,
  rounded corners=3pt,inner sep=8pt,
  drop shadow={shadow xshift=1.2pt,shadow yshift=-1.2pt,opacity=.13}]
{\begin{minipage}{\dimexpr\textwidth-18pt\relax}
\fontsize{9.2}{10.8}\selectfont
\textbf{Module-level specification}\hfill\texttt{uart\_rx}
\par\vspace{2pt}{\color{specblue!45}\rule{\linewidth}{.5pt}}\par\vspace{3pt}
\begin{minipage}[t]{.49\linewidth}
\vspace{0pt}
\textbf{Interface excerpt}\par\vspace{3pt}
\textbf{Inputs}\par
\texttt{clk\_i, rstn\_i, rx\_i}\quad(1 bit each)\par
\texttt{cfg\_div\_i[15:0], cfg\_en\_i}\par
\texttt{cfg\_bits\_i[1:0], \ldots}\par
\texttt{rx\_ready\_i, err\_clr\_i}\par\vspace{4pt}
\textbf{Outputs}\par
\texttt{rx\_data\_o[7:0]}\par
\texttt{rx\_valid\_o, busy\_o, err\_o}\par\vspace{4pt}
\textbf{Receive timing}\par
LSB first; 5--8 data bits selected by
\texttt{cfg\_bits\_i}. A full-bit timer period is
\texttt{cfg\_div\_i}+1 clocks. \ldots
\end{minipage}\hfill
\begin{minipage}[t]{.48\linewidth}
\vspace{0pt}
\textbf{Clock and reset}\par\vspace{3pt}
Rising-edge \texttt{clk\_i}; active-low asynchronous
\texttt{rstn\_i}. Reset enters idle, sets stored data to
\texttt{8'hFF}, and clears the bit counter, timer,
completion flag and sticky error. \ldots
\par\vspace{5pt}
\textbf{Output and handshake}\par\vspace{3pt}
In the save state, assert \texttt{rx\_valid\_o} and hold
\texttt{rx\_data\_o} until \texttt{rx\_ready\_i}.
There is no receive queue; the valid character is offered
before parity and stop processing. \ldots
\par\vspace{5pt}
\textbf{Priority}\par\vspace{3pt}
\texttt{err\_clr\_i} overrides setting the sticky error
on the same clock edge. \ldots
\end{minipage}
\end{minipage}};
\end{tikzpicture}
\caption{Module-level specification (Case: \texttt{uart\_rx}), including its interfaces and widths,
reset state, cycle timing, handshake, and update priority, etc.}
\label{fig:module-specification-example}
\end{minipage}
\end{figureblock}

\begin{figureblock}%
\begin{minipage}{\textwidth}
\centering
\begin{tikzpicture}
\node[draw=specteal!65,fill=specteal!4,line width=.55pt,
  rounded corners=3pt,inner sep=8pt,
  drop shadow={shadow xshift=1.2pt,shadow yshift=-1.2pt,opacity=.13}]
{\begin{minipage}{\dimexpr\textwidth-18pt\relax}
\fontsize{9.2}{10.8}\selectfont
\textbf{Project-level specification and configuration}\hfill\textbf{VGA/LCD controller}
\par\vspace{2pt}{\color{specteal!45}\rule{\linewidth}{.5pt}}\par\vspace{3pt}
\begin{minipage}[t]{.50\linewidth}
\vspace{0pt}
\textbf{Specification}\par\vspace{3pt}
\textbf{Interfaces.} 32-bit Wishbone data; 12-bit
slave addresses; 8-bit R/G/B outputs.
\texttt{wb\_clk\_i} drives the bus domain and
\texttt{clk\_p\_i} the pixel domain. \ldots
\par\vspace{5pt}
\textbf{Reset.} \texttt{rst\_i} is asynchronous and
active low; \texttt{wb\_rst\_i} is held at 0.
No startup RAM contents may be assumed. \ldots
\par\vspace{5pt}
\textbf{Behavior.} Program timing and frame-buffer
base addresses through Wishbone registers. Fetch
packed pixels most-significant first; preserve
transactions while ACK is delayed. FIFO underrun
sets \texttt{STAT[1]} and raises an interrupt. \ldots
\par\vspace{5pt}
\textbf{Hierarchy.} 15 required RTL files/modules:
Wishbone slave/master, timing and pixel generators,
line FIFOs, generic RAMs, \ldots
\end{minipage}\hfill
\begin{minipage}[t]{.47\linewidth}
\vspace{0pt}
\textbf{Fixed configuration}\par\vspace{3pt}
\begin{tabular}{@{}l@{\quad}l@{}}
\multicolumn{2}{@{}l@{}}{Case: \texttt{opencores\_vga\_rgb\_v1}}\\
Top & \texttt{vga\_enh\_top}\\[3pt]
\texttt{ARST\_LVL} & \texttt{0}\\
\texttt{LINE\_FIFO\_AWIDTH} & \texttt{7}\\[3pt]
Output & Parallel RGB888\\
Optional DVI/BIST & Disabled\\
RAM implementation & Generic\\[3pt]
Wishbone clock & 4.8\,ns period\\
Pixel clock & 30\,ns period\\
\end{tabular}
\end{minipage}
\end{minipage}};
\end{tikzpicture}
\caption{Project-level specification paired with the configuration for one RTL reference.}
\label{fig:project-specification-example}
\end{minipage}
\end{figureblock}
\endgroup

\section{Complete Task Definitions}
\label{app:tasks}

Table~\ref{tab:task-contracts} summarizes 12 task with their indexing. We clearly define the task formulation here, including the input, output, and verification methods. Direct tasks use supplied specifications or issue reports; Round-trip tasks first derive an intermediate specification from clean RTL and then use it in a separate recovery stage. Reference RTL and verification assets remain hidden from the evaluated generation, completion, or repair stage.

\begin{table}[htbp]
  \centering
  \caption{Task contracts in \systemname. Arrows indicate input--output mappings; parentheses specify verification methods.}
  \label{tab:task-contracts}
  \scriptsize
  \setlength{\tabcolsep}{6pt}
  \renewcommand{\arraystretch}{1.08}
  \begin{tabular}{@{}c>{\raggedright\arraybackslash}p{5.1cm}>{\raggedright\arraybackslash}p{5.7cm}@{}}
    \toprule
    \textbf{Task} & \textbf{Task name} & \textbf{Input $\to$ output (verification)} \\
    \midrule
    1.1 & \taskname{1.1} & Specification $\to$ module (formal verification). \\
    1.2 & \taskname{1.2} & Specification $\to$ project RTL (testbench or formal verification). \\
    1.3 & \taskname{1.3} & Module $\to$ specification $\to$ module (formal verification). \\
    1.4 & \taskname{1.4} & Project RTL $\to$ specification $\to$ project RTL (testbench or formal verification). \\
    \midrule
    2.1 & \taskname{2.1} & Specification + masked module $\to$ completion (formal verification). \\
    2.2 & \taskname{2.2} & Specification + retained project $\to$ missing module (testbench or formal verification). \\
    2.3 & \taskname{2.3} & Module $\to$ specification; specification + masked module $\to$ completion (formal verification). \\
    2.4 & \taskname{2.4} & Project RTL $\to$ specification; specification + retained project $\to$ completion (testbench or formal verification). \\
    \midrule
    3.1 & \taskname{3.1} & Specification + module mutant $\to$ repair (formal verification). \\
    3.2a & \taskname{3.2a} & Nonfunctional issue + project RTL $\to$ patch (commit match). \\
    3.2b & \taskname{3.2b} & Functional issue + faulty project RTL $\to$ patch (formal verification). \\
    3.2c & \taskname{3.2c} & Specification + project mutant $\to$ patch (formal verification). \\
    3.3 & \taskname{3.3} & Module $\to$ specification; specification + mutant $\to$ repair (formal verification). \\
    3.4 & \taskname{3.4} & Project RTL $\to$ specification; specification + mutant $\to$ patch (formal verification). \\
    \bottomrule
  \end{tabular}
\end{table}

\textbf{Notation and scoring.}
Let $f_\theta$ denote the model or agent, $D$ a clean module, $P$ a clean project, $S$ a supplied specification, and $A$ a derived specification. Write $X_{\setminus R}$ for RTL with region $R$ removed and $X\oplus Y$ for inserting a completion or applying a patch. The formal score $F(X,R)$ is one when $X$ is proven equivalent to reference $R$, and zero otherwise. For project generation and completion, $V_P(X)$ uses the case's declared verification method: the fraction of frozen testbenches passed, or the binary formal score. Each complete testbench is one testpoint. Invalid RTL receives zero.

\textbf{Round-trip constraints.}
Specification extraction and recovery use isolated contexts; only the validated intermediate specification crosses between stages. Each case declares its permitted representation and length limit, and rejects copied RTL. A failed intermediate check gives an end-to-end score of zero without recovery. The formulas below describe recovery after this check passes.

\begin{samepage}
\subsection{Type 1: Generation}

\paragraph{Task 1.1: \taskname{1.1}.}
The model generates a complete module from a validated specification. Formal verification compares the generated RTL with the hidden clean module.
\begin{equation}
  D'=f_\theta(S), \qquad s_{1.1}=F(D',D).
  \label{eq:derived-spec}
\end{equation}
\end{samepage}

\paragraph{Task 1.2: \taskname{1.2}.}
The agent receives a project specification, declared interfaces, and permitted public assets, then implements the required project RTL. Hidden testbenches or formal verification evaluate the implementation.
\begin{equation}
  P'=f_\theta(S_P), \qquad s_{1.2}=V_P(P').
  \label{eq:single-target}
\end{equation}

\begin{figure}[htbp]
  \centering
  \includegraphics[width=0.52\linewidth]{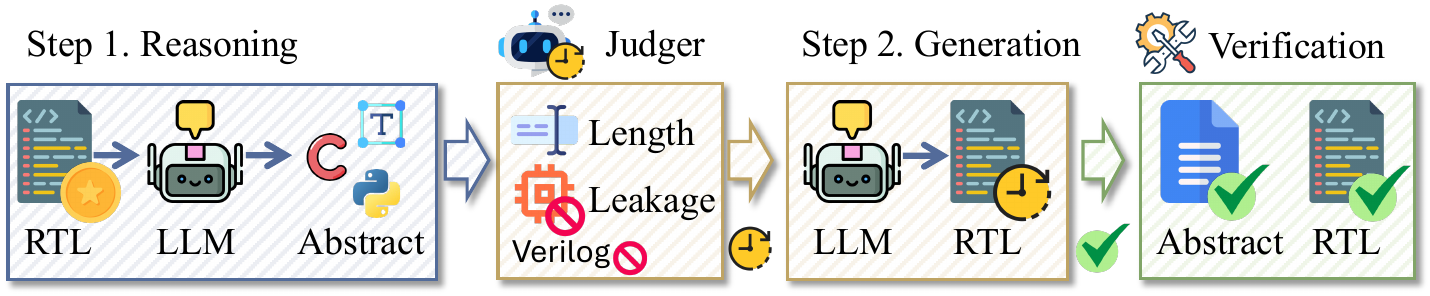}
  \caption{Task~1.3 reconstructs a module from an intermediate specification without access to the original RTL.}
  \label{fig:round-trip}
\end{figure}

\paragraph{Task 1.3: \taskname{1.3}.}
The first stage describes a clean module; a fresh call reconstructs it from that description alone (Figure~\ref{fig:round-trip}). Intermediate representations include natural language, diagram-oriented descriptions, executable Python or C, and free-form abstractions, subject to the declared format and length limits.
\begin{equation}
  A=f_\theta(D), \qquad D'=f_\theta(A), \qquad s_{1.3}=F(D',D).
  \label{eq:round-trip}
\end{equation}

\paragraph{Task 1.4: \taskname{1.4}.}
The first stage derives a specification from the clean project. A fresh agent receives this specification and the public interface contract, then reconstructs the required project without the reference RTL.
\begin{equation}
  A_P=f_\theta(P), \qquad P'=f_\theta(A_P), \qquad s_{1.4}=V_P(P').
  \label{eq:project-round-trip}
\end{equation}

\subsection{Type 2: Completion}

\paragraph{Task 2.1: \taskname{2.1}.}
The model receives a supplied specification and a module with a missing RTL region $B$. It reconstructs that region, and formal verification checks the completed module.
\begin{equation}
  B'=f_\theta(S,D_{\setminus B}), \qquad
  s_{2.1}=F(D_{\setminus B}\oplus B',D).
  \label{eq:direct-module-completion}
\end{equation}

\paragraph{Task 2.2: \taskname{2.2}.}
The agent implements a missing module $M$ from its validated specification and the retained project context. The recovered module is reinserted into the project for hidden evaluation.
\begin{equation}
  M'=f_\theta(S_M,P_{\setminus M}), \qquad
  s_{2.2}=V_P(P_{\setminus M}\oplus M').
  \label{eq:spec-project-mask}
\end{equation}

\begin{figure}[htbp]
  \centering
  \includegraphics[width=0.52\linewidth]{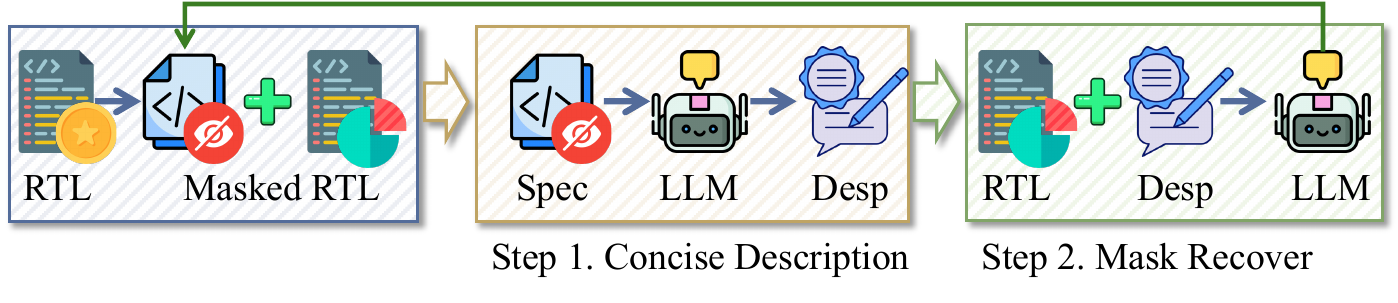}
  \caption{Task~2.3 reconstructs masked RTL using derived intent and the retained module context.}
  \label{fig:mask-task}
\end{figure}

\paragraph{Task 2.3: \taskname{2.3}.}
The first stage derives module intent from clean RTL. A fresh call receives this intent and the masked module, then reconstructs the missing block (Figure~\ref{fig:mask-task}). Masks cover complete syntactic regions, such as \texttt{always}, \texttt{assign}, or \texttt{case} blocks.
\begin{equation}
  A=f_\theta(D), \qquad B'=f_\theta(A,D_{\setminus B}), \qquad
  s_{2.3}=F(D_{\setminus B}\oplus B',D).
  \label{eq:masked}
\end{equation}

\paragraph{Task 2.4: \taskname{2.4}.}
The first stage describes the missing region $R$ in its clean project context. A fresh agent uses this description and the retained project to reconstruct $R$, which may be a logic region or a module. Evaluation checks the reintegrated project.
\begin{equation}
  A_R=f_\theta(R,P), \qquad R'=f_\theta(A_R,P_{\setminus R}), \qquad
  s_{2.4}=V_P(P_{\setminus R}\oplus R').
  \label{eq:project-mask}
\end{equation}

\subsection{Type 3: Debugging}

Direct Project-level debugging is Task~3.2, with three scored subtypes: nonfunctional issue fixing (3.2a), functional issue fixing (3.2b), and mutant repair (3.2c).

Let $\mu(X)$ denote a compile-valid mutant whose behavior differs from the clean reference, $I$ an issue report, and $\Delta$ a predicted patch. Easy mode supplies the reviewed fault location; hard mode requires correct identification before repair. Below, $g=1$ in easy mode and is the identification-pass indicator in hard mode; mode-specific location inputs are omitted for brevity. Failed identification stops repair and yields zero.

\paragraph{Task 3.1: \taskname{3.1}.}
The model repairs a module mutant using the original validated specification. Formal verification compares the repaired module with its clean parent.
\begin{equation}
  \Delta=f_\theta(S,\mu(D)), \qquad
  s_{3.1}=g\,F(\mu(D)\oplus\Delta,D).
  \label{eq:spec-debug}
\end{equation}

\paragraph{Task 3.2a: \taskname{3.2a}.}
The agent addresses a nonfunctional issue in a buildable project while preserving its behavior. The predicted patch is evaluated by commit match against the withheld developer change $\Delta^\star$.
\begin{equation}
  \Delta=f_\theta(I_{\mathrm{nonfunc}},P), \qquad
  s_{3.2a}=g\,\operatorname{CommitMatch}(\Delta,\Delta^\star).
  \label{eq:nonfunctional-issue-repair}
\end{equation}

\begin{figure}[htbp]
  \centering
  \includegraphics[width=0.52\linewidth]{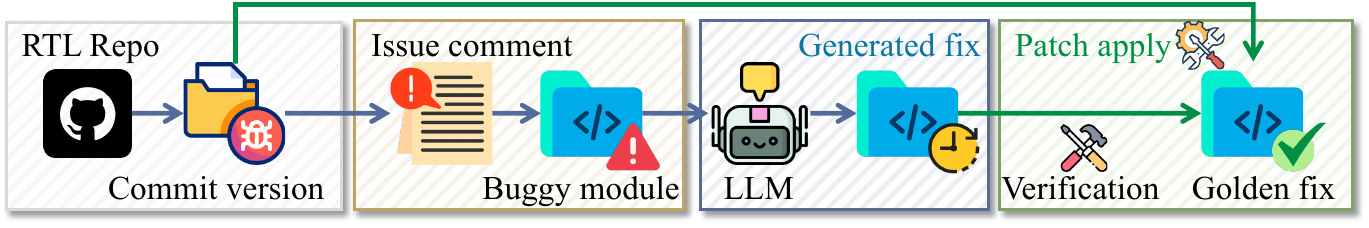}
  \caption{Task~3.2b repairs a repository issue and verifies the result against the withheld developer-fixed behavior.}
  \label{fig:repo-task}
\end{figure}

\paragraph{Task 3.2b: \taskname{3.2b}.}
The agent receives a functional issue and the buildable buggy revision $P_{\mathrm{bug}}$. Formal verification compares the repaired RTL with the developer-fixed reference $P_{\mathrm{fix}}$ (Figure~\ref{fig:repo-task}).
\begin{equation}
  \Delta=f_\theta(I_{\mathrm{func}},P_{\mathrm{bug}}), \qquad
  s_{3.2b}=g\,F(P_{\mathrm{bug}}\oplus\Delta,P_{\mathrm{fix}}).
  \label{eq:issue-repair}
\end{equation}

\paragraph{Task 3.2c: \taskname{3.2c}.}
The agent repairs a project mutant using the unchanged original project specification. Formal verification checks the repaired design against the clean parent; mutation metadata and the golden repair remain hidden.
\begin{equation}
  \Delta=f_\theta(S_P,\mu(P)), \qquad
  s_{3.2c}=g\,F(\mu(P)\oplus\Delta,P).
  \label{eq:spec-mutant-repair}
\end{equation}

\paragraph{Task 3.3: \taskname{3.3}.}
The first stage derives intent from the clean module. A fresh call repairs its mutant using only that intent and the faulty RTL. Formal verification checks equivalence to the clean module.
\begin{equation}
  A=f_\theta(D), \qquad \Delta=f_\theta(A,\mu(D)), \qquad
  s_{3.3}=g\,F(\mu(D)\oplus\Delta,D).
  \label{eq:round-debug}
\end{equation}

\paragraph{Task 3.4: \taskname{3.4}.}
The first stage derives a specification from the clean project. A fresh agent uses it to repair the mutant project, with formal verification against the hidden clean reference.
\begin{equation}
  A_P=f_\theta(P), \qquad \Delta=f_\theta(A_P,\mu(P)), \qquad
  s_{3.4}=g\,F(\mu(P)\oplus\Delta,P).
  \label{eq:mutant-repair}
\end{equation}

\section{Additional Benchmarking Results}
\label{app:additional-results}

\subsection{Data Contamination Analysis}\label{sec:data-contamination-analysis}
\begingroup

\newcommand{\SensitivityPanelAbove}{3pt}%
\newcommand{\SensitivityPanelBelow}{5pt}%
\newcommand{\SensitivityTableCaptionGap}{3pt}%
\newcommand{\SensitivityFigureCaptionGap}{2pt}%
\newcommand{\SensitivityPanelWidth}{0.94\linewidth}
\newcommand{\SensitivityFigureScale}{0.96}%
\newcommand{\SensitivityTableFont}{\fontsize{9}{10.5}\selectfont}
\newcommand{\SensitivityCaptionFont}{\fontsize{9}{10.5}\selectfont}

\definecolor{deltapositive}{RGB}{0,128,64}
\definecolor{deltanegative}{RGB}{190,35,40}
\makeatletter
\edef\@currentlabel{\thesubsection}
\def\@currentlabelname{Data Contamination Analysis}
\label{sec:specification-sensitivity}
\providecommand{\tablecaption}[1]{\def\@captype{table}\caption{#1}}
\def\@captype{figure}
\let\SensitivityOriginalMakeCaption\@makecaption
\long\def\@makecaption#1#2{{%
  \SensitivityCaptionFont\SensitivityOriginalMakeCaption{#1}{#2}}}
\makeatother
\setlength{\abovecaptionskip}{0pt}
\setlength{\belowcaptionskip}{0pt}

Following the overfitting detection methodology introduced in RTL-BenchMT~\citep{wang2026rtl}, we further conduct data contamination analysis on \emph{Direct
Module-level Generation} task cases. Specifically, we rewrite one variant for each specification. Each specification variant uses a different description style but preserves all semantics of the original. Figure~\ref{fig:salv-verilogeval-sensitivity}
shows an illustration based on the QiMeng-SALV in VerilogEval~\cite{liu2023verilogeval}. In this matrix, \emph{FN} means that the model passes the original specification while failing the variant. We take \emph{FN} as an indicator of overfitting from data contamination on the benchmark cases.
Table~\ref{tab:one-variant-benchmark-comparison} summarizes the measured
accuracy changes ($\Delta$) and FN rates. If overall performance decreases ($\Delta < 0$), the model may also be at risk of data contamination. From the table, our current benchmark cases have relatively low contamination risk.

\par\vspace{\SensitivityPanelAbove}
\noindent\makebox[\linewidth][c]{%
\begin{minipage}{\SensitivityPanelWidth}
\begin{minipage}[t]{0.64\linewidth}
\vspace{0pt}
\tablecaption{Overfitting analysis on the direction RTL generation task.}
\label{tab:one-variant-benchmark-comparison}
\par\vspace{\SensitivityTableCaptionGap}
\centering\SensitivityTableFont
\setlength{\tabcolsep}{2.5pt}
\renewcommand{\arraystretch}{1.00}
\begin{tabular*}{\linewidth}{@{\extracolsep{\fill}}llrr@{}}
\toprule
Benchmark & Model & $\Delta$ (\%)$\uparrow$ & FN (\%)$\downarrow$ \\
\midrule
\multirow{3}{*}{RTLLM} & GPT-5.4 & \textcolor{deltapositive}{\(+2.00\)} & 4.00 \\
 & CodeV-R1 & \textcolor{deltapositive}{\(+6.00\)} & \textbf{4.00} \\
 & QiMeng-SALV & \textcolor{deltanegative}{\(-2.00\)} & 6.00 \\
\midrule
\multirow{3}{*}{VerilogEval} & GPT-5.4 & \textcolor{deltapositive}{\(+1.28\)} & 3.85 \\
 & CodeV-R1 & \textcolor{deltanegative}{\(-0.64\)} & 5.77 \\
 & QiMeng-SALV & \textcolor{deltanegative}{\(-0.64\)} & 6.41 \\
\midrule
\multirow{3}{*}{RTL-BenchLS} & GPT-5.4 & \textcolor{deltanegative}{\(-1.29\)} & \textbf{2.76} \\
 & CodeV-R1 & \textcolor{deltapositive}{\(+5.52\)} & 4.24 \\
 & QiMeng-SALV & \textcolor{deltapositive}{\(+3.68\)} & \textbf{3.68} \\
\bottomrule\end{tabular*}

\end{minipage}\hfill
\begin{minipage}[t]{0.32\linewidth}
\vspace{0pt}
\centering
\includegraphics[width=\SensitivityFigureScale\linewidth,
  trim=0 10bp 0 3bp,clip]{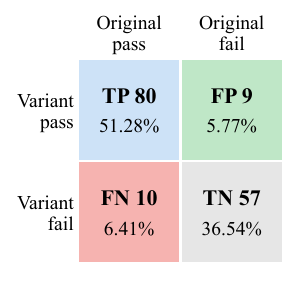}
\par\vspace{\SensitivityFigureCaptionGap}
\caption{Overfitting detection matrix, illustrated by QiMeng-SALV on VerilogEval.}
\label{fig:salv-verilogeval-sensitivity}
\end{minipage}
\end{minipage}}
\par\vspace{\SensitivityPanelBelow}

\endgroup

\subsection{Round-Trip Reasoning Analysis}\label{sec:round-trip-quality}
\begingroup
\newcommand{\RoundTripBarPlotAbove}{2pt}
\newcommand{\RoundTripBarPlotBelow}{4pt}
\newcommand{\RoundTripBarPlotCaptionGap}{2pt}
\makeatletter
\edef\@currentlabel{\thesubsection}
\def\@currentlabelname{Round-Trip Reasoning Analysis}
\label{sec:roundtrip-intermediate-analysis}
\def\@captype{figure}
\let\RoundTripOriginalMakeCaption\@makecaption
\long\def\@makecaption#1#2{{%
  \small\RoundTripOriginalMakeCaption{#1}{#2}}}
\makeatother
\setlength{\abovecaptionskip}{0pt}
\setlength{\belowcaptionskip}{0pt}

The round-trip reasoning is challenging and involves two steps. The quality of the intermediate representation in step (1) may significantly influence the performance in step (2). To more clearly quantify the quality of the intermediate representations, we define the \textbf{Requirement Fidelity Score (RFS)}.
Specifically, we first extract a checklist
of requirements for a qualified intermediate representation, such as necessary
clock/reset behavior, state conditions, etc. Then we use the strongest commercial model, such as GPT-6.1 Sol, to check the intermediate representation against this checklist. Specifically, each
requirement is labeled as preserved (scored 1), partial(scored 0.5), omitted (scored 0), or contradicted (scored -1). We calculate the \textbf{Coverage} score by summing the weighted score from preserved and partial labels, and the \textbf{Contradiction} score by summing the contradicted labels.
Let $N$ be the checklist size, and $n_f$, $n_p$ and $n_c$ the numbers of fully
preserved, partially preserved and contradicted requirements. The scores are defined as follows.
\[
\mathrm{Coverage}=100\frac{n_f+0.5n_p}{N},\qquad
\mathrm{Contradiction}=100\frac{n_c}{N}.
\]

Finally, the \textbf{Requirement Fidelity Score (RFS)} is calculated from \emph{Coverage} and \emph{Contradiction}: \textbf{RFS = Coverage $-$ Contradiction},
ranged in $[-100,100]$, with higher score denoting the better quality.

\par\vspace{\RoundTripBarPlotAbove}
\noindent\begin{minipage}{\linewidth}
\centering
\includegraphics[width=1.0\textwidth]{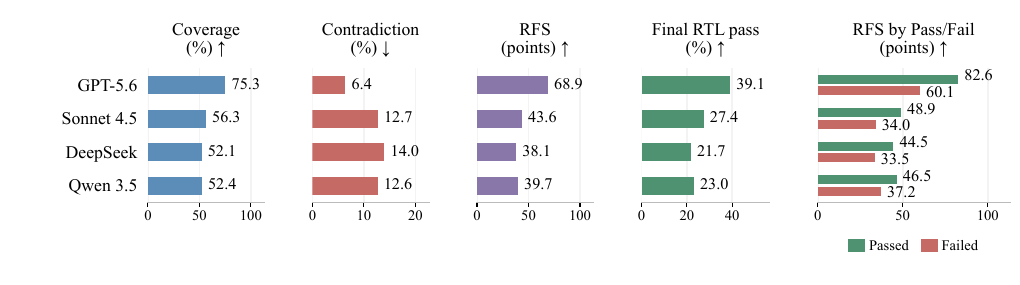}
\par\vspace{\RoundTripBarPlotCaptionGap}
\caption{Intermediate-description quality and final RTL functional
pass rates. Arrows mark preferred directions.}
\label{fig:roundtrip-intermediate-and-final-pass}
\end{minipage}
\par\vspace{\RoundTripBarPlotBelow}

Figure~\ref{fig:roundtrip-intermediate-and-final-pass} analysis score for the intermediate evaluation score based on slice 01 at \emph{Direct Module-level RTL Generation} task. The figure indicates that higher quality of intermediate representations (higher RFS score) contributes to higher final RTL pass rates. Additionally, we calculate the RFS score based on PASS and Fail groups (final RTL pass results). The result shows that passed RTL deisgns generally benefit from higher intermediate representation quality. Although we observed that most intermediate representations cannot achieve 100 RFS or 100\% coverage, the LLMs can still generate correct RTL designs if the representation preserves enough quality.
\endgroup

\clearpage
\subsection{Token Consumption and Effective Project Generation}
\label{sec:token-consumption}

Figure~\ref{fig:token-consumption} shows token consumption for functional outcomes for Direct Project-level Design Generation (Task~1.2) within one repetition ($n=1$). For DeepSeek, the averaged token consumption rises from 684k tokens on Simple projects to 965k on Hard projects, while the mean testpoint pass rate falls from 33.8\% to 0\%. On Medium projects, approximately 92\% of its recorded tokens are spent on attempts that pass no testpoints; on Hard projects, all tokens failed. Llama consumes fewer tokens (95k--229k per project) but passes no testpoints in any group, so lower consumption alone does not indicate effective generation. These results show that \emph{current models are limited on project-level RTL tasks, with substantial token consumption that yields much less functional pass.}

\begin{figureblock}%
  \begin{minipage}{\textwidth}
  \centering
  \includegraphics[width=\textwidth]{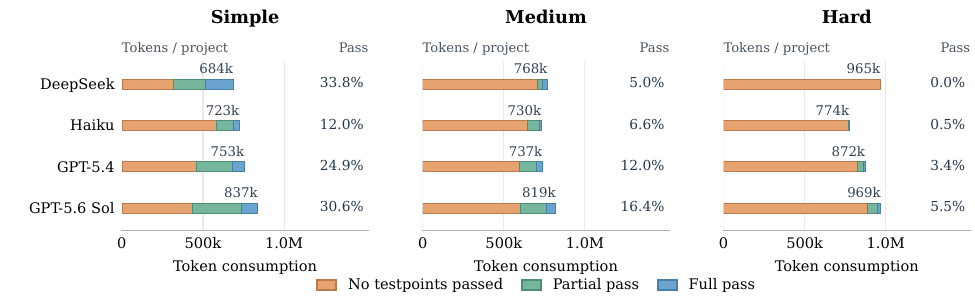}
  \caption{Token consumption and functional outcomes for Task~1.2. Bars show input plus output tokens per project, partitioned by final outcome and averaged over all projects in each group; percentages are mean per-project testpoint pass rates.}
  \label{fig:token-consumption}
  \end{minipage}
\end{figureblock}

\clearpage
\setlength{\intextsep}{6pt}
\subsection{Module-level round-trip generation with different formats}
\label{app:representation-analysis}

\begin{tableblock}%
  \begin{minipage}{\textwidth}
  \centering
  \caption{Task~1.3 functional pass rates (\%).}
  \label{tab:additional-task1}
  \resizebox{\textwidth}{!}{%
  \begin{tabular}{llrrrrrrr}
    \toprule
    \textbf{Representation} & \textbf{Setting} & \textbf{GPT-5.4} & \textbf{Sonnet} & \textbf{Haiku} & \textbf{DeepSeek} & \textbf{Mistral} & \textbf{Qwen} & \textbf{LLaMA} \\
    \midrule
    $\mathcal{A}_{\mathrm{NL}}$ & $r=0.5$ & $35.3$ & 17.1 & 11.7 & 10.3 & 6.8 & 4.6 & 7.9 \\
    $\mathcal{A}_{\mathrm{NL\text{-}Diag}}$ & $r\approx0.9$ & $46.1$ & 39.5 & 30.3 & 16.4 & 24.6 & 3.7 & 3.3 \\
    $\mathcal{A}_{\mathrm{Code}}$ & Python ($r\approx0.9$) & $49.7$ & 52.1 & 31.4 & 39.5 & 31.4 & 0.7 & 18.8 \\
    $\mathcal{A}_{\mathrm{Code}}$ & C ($r\approx1.1$) & $43.2$ & 39.5 & 26.4 & 33.8 & 28.3 & 28.3 & 19.8 \\
    $\mathcal{A}_{\mathrm{Free}}$ & $r=0.5$ & $38.0$ & 20.6 & 16.0 & 14.1 & 19.2 & 15.0 & 14.8 \\
    $\mathcal{A}_{\mathrm{Free}}$ & $r=1.0$ & $48.8$ & 34.0 & 27.4 & 22.7  & 28.9 & 26.9 & 18.4 \\
    \bottomrule
  \end{tabular}}
  \end{minipage}
\end{tableblock}

\begin{figureblock}%
  \begin{minipage}{\textwidth}
  \centering
  \includegraphics[width=\textwidth]{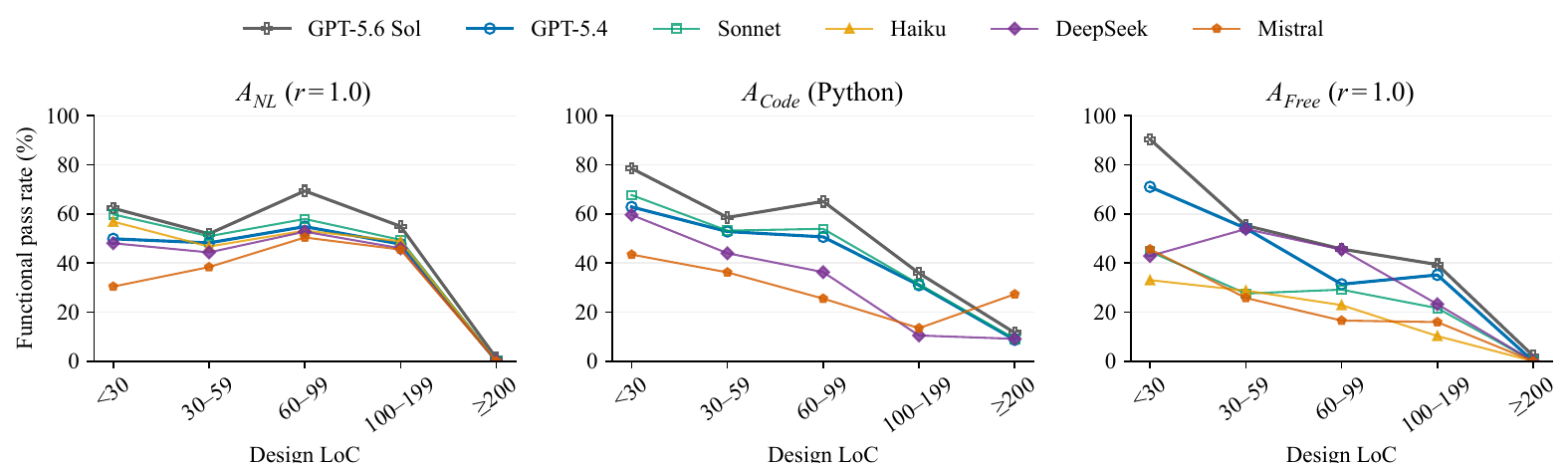}
  \caption{Results on Round-trip module-level generation tasks.}
  \label{fig:loc-task13-variants}
  \end{minipage}
\end{figureblock}

For round-trip tasks, we can explicitly require the format for intermediate representations. For example, in addition to the natural language ($\mathcal{A}_\text{NL}$), we can also require the agent to generate Diagram-oriented ($\mathcal{A}_\text{NL-Diag}$) and executable code ($\mathcal{A}_\text{NL-code}$).

\paragraph{Executable code as intermediate representation.} For \emph{module-level round-trip RTL generation} tasks, an interesting exploration is to utilize software code as a representation. We design these experiments to examine whether the pre-trained knowledge on software code can help with hardware design. 
We pick Python and C as two representative code-based intermediate representations. Python has usually helped achieve higher final RTL pass rates for Sonnet, Haiku, DeepSeek, and Mistral. On the contrary, Qwen3.5 and Llama perform better with C. We observe that Python code as an intermediate representation performs clearly better than other intermediate representations.

\paragraph{Executable intermediate versus final RTL.} Because executable code can also generate simulation traces. We compare the software code's simulation results against the reference RTL.
Figure~\ref{fig:intermediate-vs-final} demonstrates the results. A considerable proportion of generated software code simulates equivalently to the reference RTL. The final RTL can outperform its intermediate on the simulation-based check. This phenomenon implies that an imperfect intermediate may still preserve enough behavioral information for reconstruction.

\begin{figureblock}%
  \begin{minipage}{\textwidth}
  \centering
  \includegraphics[width=\linewidth]{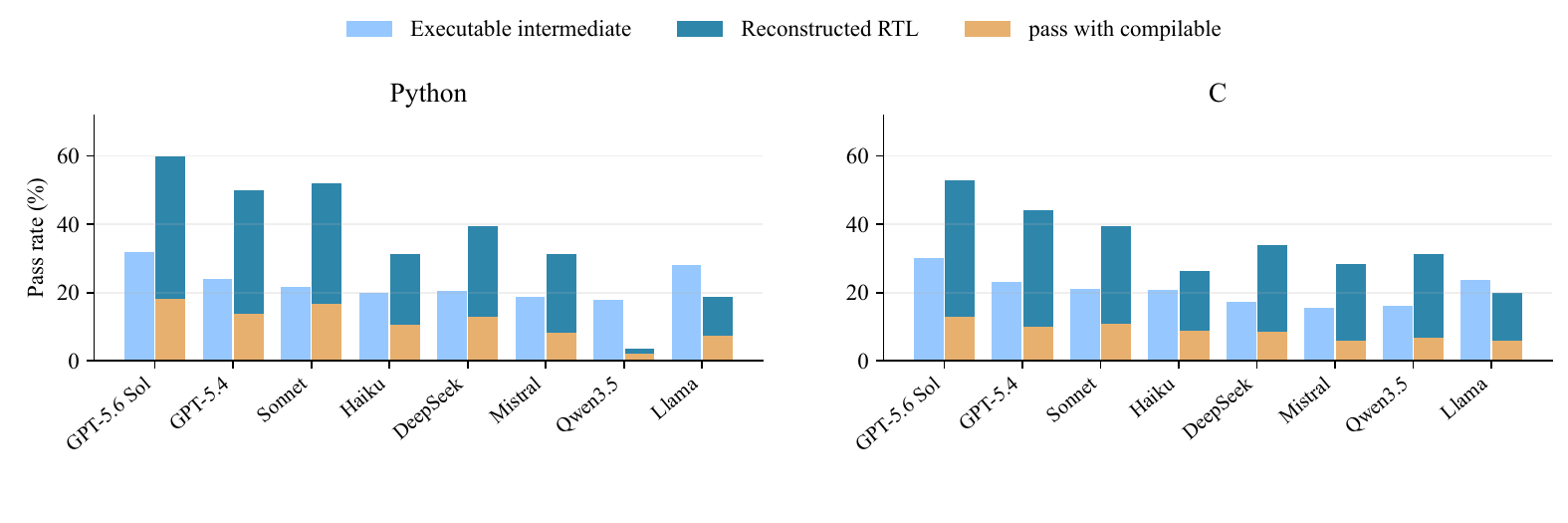}
  \caption{Task~1.3 executable-intermediate and reconstructed-RTL pass
  rates for Python and C.}
  \label{fig:intermediate-vs-final}
  \end{minipage}
\end{figureblock}

\setlength{\intextsep}{12pt plus 2pt minus 2pt}

\clearpage
\subsection{Token Consumption in Task~3.2 Issue Fixing}
\label{sec:issue-token-consumption}

\paragraph{Direct Project-level debugging subtype scores.}
Table~\ref{tab:direct-project-debugging-subtypes} reports the detailed debugging results on three subtypes. Figure~\ref{fig:issue-token-consumption} separates token consumption by final outcome for the issue-fixing subtypes of Task~3.2: nonfunctional issues (3.2a) and functional issues (3.2b). Easy mode supplies the issue location and distinguishes failed from successful repairs. Hard mode additionally distinguishes unsuccessful identification from successful identification followed by a failed repair. From the token consumption analysis, we observe that issue fixing is harder than issue identification. Many tokens succeed on the issue identification step while failing at the fixing step.

\begin{tableblock}%
  \begin{minipage}{\textwidth}
  \centering
  \caption{Direct Project-level debugging subtype functional scores (\%).}
  \label{tab:direct-project-debugging-subtypes}
  \resizebox{\textwidth}{!}{%
  \begin{tabular}{lrrrrrrrr}
    \toprule
    \textbf{Subtype} & \shortstack{\textbf{GPT}\\\textbf{5.6 Sol}} & \shortstack{\textbf{GPT}\\\textbf{5.4}} & \shortstack{\textbf{Sonnet}\\\textbf{4.5}} & \shortstack{\textbf{Haiku}\\\textbf{4.5}} & \shortstack{\textbf{DeepSeek}\\\textbf{V3.2}} & \shortstack{\textbf{Mistral}\\\textbf{Large}} & \shortstack{\textbf{Qwen}\\\textbf{3.5}} & \shortstack{\textbf{Llama}\\\textbf{3.3}} \\
    \midrule
    Nonfunctional (CommitMatch) & 44.6 & 34.7 & 39.6 & 30.0 & 31.7 & 19.8 & 24.8 & 14.9 \\
    Nonfunctional (Checklist) & 50.1 & 41.2 & 45.6 & 37.0 & 38.5 & 27.8 & 32.3 & 23.4 \\
    Functional issue fixing & 10.0 & 7.0 & 5.6 & 3.7 & 12.0 & 6.5 & 4.6 & 5.3 \\
    Mutant debugging & 48.8 & 61.2 & 67.5 & 63.8 & 40.8 & 42.7 & 20.8 & 9.9 \\
    \bottomrule
  \end{tabular}}
  \end{minipage}
\end{tableblock}

\begin{figureblock}%
  \begin{minipage}{\textwidth}
  \centering
  \includegraphics[width=\textwidth]{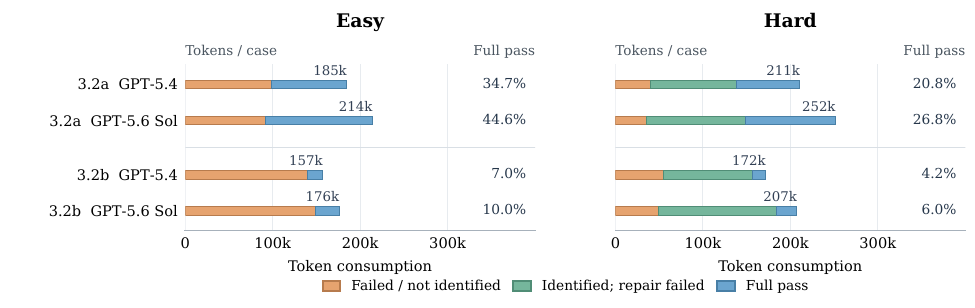}
  \caption{Issue-fixing token consumption and pass rates for Task~3.2 subtypes 3.2a and 3.2b.}
  \label{fig:issue-token-consumption}
  \end{minipage}
\end{figureblock}

\section{Slice Balance and Additional-Slice Evaluation}
\label{app:slice-representativeness}

In this section, we report the audit of the slices 01, 05, 10, 15, and 20 for round-trip
module-level tasks to demonstrate that our slicing machenism is balanced. This ensures that a single slice's result can represent the whole round-trip dataset on module-level tasks.

\begin{figure}[!htb]
  \centering
  \includegraphics[width=\textwidth]{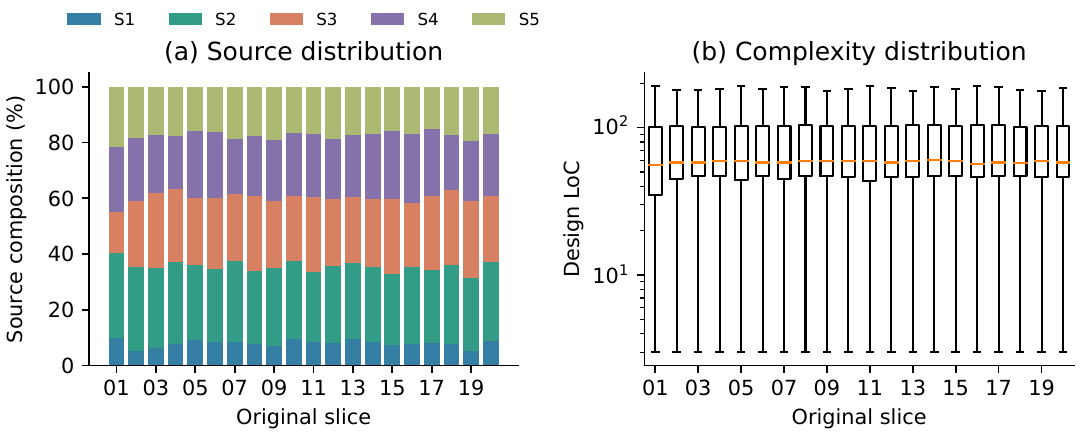}
  \caption{Balance of the 20 original released slices. Left: S1--S5 source
  composition. Right: design LoC distributions on a logarithmic axis;
  boxes show the interquartile range and median, with outliers omitted from
  display.}
  \label{fig:original-slice-balance}
\end{figure}

\begin{figure}[!htb]
  \centering
  \includegraphics[width=\textwidth]{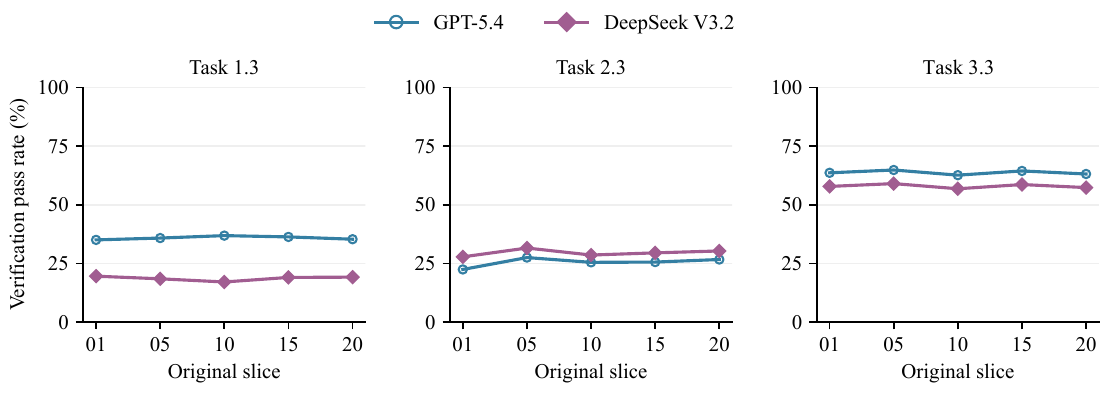}
  \caption{Five-slice verification pass rates for round-trip module-level
  generation (Task~1.3), completion (Task~2.3), and debugging (Task~3.3).}
  \label{fig:multislice-performance}
\end{figure}

\begin{figure}[!htb]
  \centering
  \includegraphics[width=\textwidth]{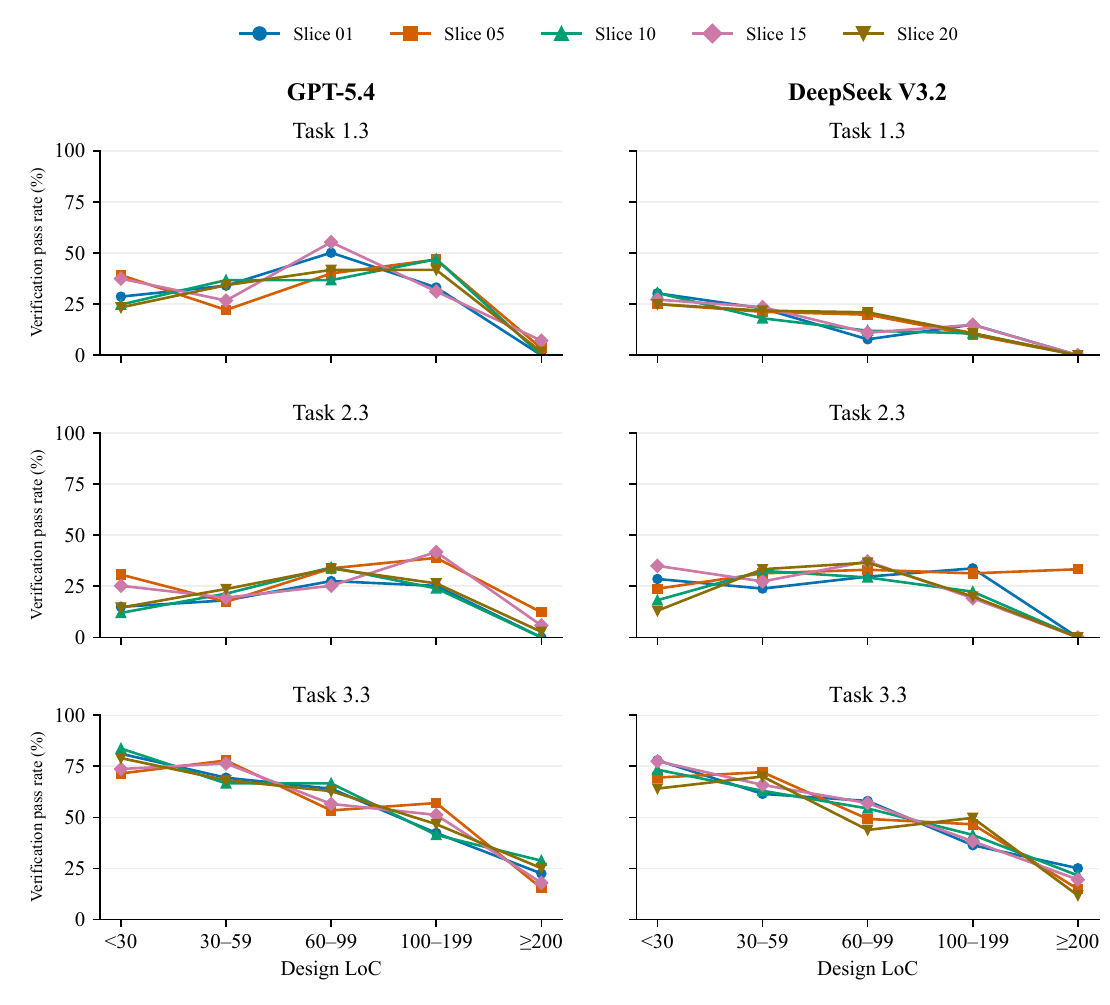}
  \caption{LoC-conditioned verification pass rates by model and slice.
  Columns: GPT-5.4 (left), DeepSeek V3.2 (right). Rows: Tasks~1.3, 2.3,
  and 3.3. Each panel shows slices~01, 05, 10, 15, and 20.}
  \label{fig:multislice-loc-control}
\end{figure}

Figure~\ref{fig:original-slice-balance} shows source composition and design
size across all 20 original released slices. 
Mean design LoC ranges from 70.1 to 74.1 across slices, and median LoC ranges
from 56 to 60. Source composition also remains broadly aligned across slices.
This demonstrates that the slicing is balanced.
Figure~\ref{fig:multislice-performance} compares slice-level performance; GPT-5.4 and DeepSeek remain within a stable performance range across slices. 
Figure~\ref{fig:multislice-loc-control} shows GPT-5.4 and DeepSeek performance as the design size increases. The results demonstrate that different models demonstrate similar performance trends on each slice.

\clearpage
\section{Detailed Analysis and Case Studies}
\label{app:analysis}

\subsection{Representative Cases for Module-level Round-Trip Tasks}
\label{app:cases}

\begin{figureblock}
  \begin{minipage}[t]{0.52\linewidth}
    \vspace{0pt}
    \centering
    \includegraphics[width=\linewidth]{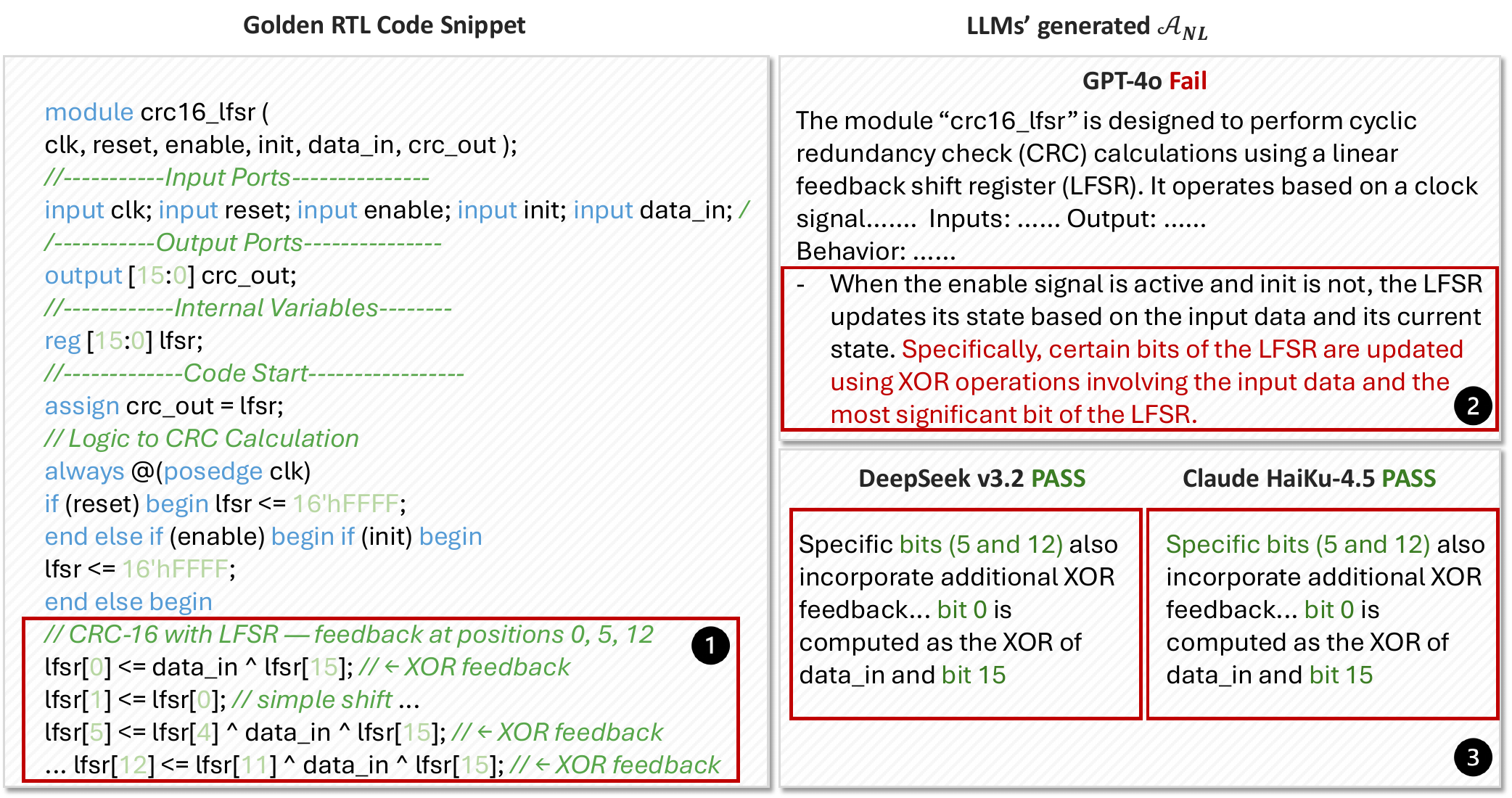}
    \caption{Task~1.3 case: the abstract must preserve the exact LFSR feedback positions to reconstruct the CRC behavior.}
    \label{fig:case-crc}
  \end{minipage}
  \hfill
  \begin{minipage}[t]{0.44\linewidth}
    \vspace{0pt}
    \textbf{Case 1: round-trip design generation.}\par\smallskip
    Figure~\ref{fig:case-crc} shows a 49-LoC, 16-bit CRC design. The reference uses LFSR feedback XORs at bit positions 0, 5, and 12. GPT-4o produces a vague intermediate representation stating only that ``certain bits'' use XOR, so the reconstruction loses the exact taps and fails equivalence. DeepSeek-v3.2 and Haiku-4.5 explicitly preserve all three indices and reconstruct equivalent designs. The case illustrates how a superficially plausible specification can omit one low-level detail that determines the complete behavior.
  \end{minipage}
\end{figureblock}

\begin{figureblock}
  \begin{minipage}[t]{0.44\linewidth}
    \vspace{0pt}
    \textbf{Case 2: round-trip completion.}\par\smallskip
    Figure~\ref{fig:case-mask} masks the overflow logic of an 8-bit add/subtract design. The golden block sets \texttt{overflow} to the XOR of \texttt{intermediate\_carry[6]} and \texttt{carry}, covering both paths. DeepSeek-v3.2 and Mistral Large describe overflow in terms of carry XOR and reconstruct the correct formula. GPT-4o and Sonnet-4.5 interpret it as generic signed overflow and emit a familiar sign-bit formula that fails on subtraction. The surrounding design is not sufficient unless the model infers the local convention precisely.
  \end{minipage}
  \hfill
  \begin{minipage}[t]{0.52\linewidth}
    \vspace{0pt}
    \centering
    \includegraphics[width=\linewidth]{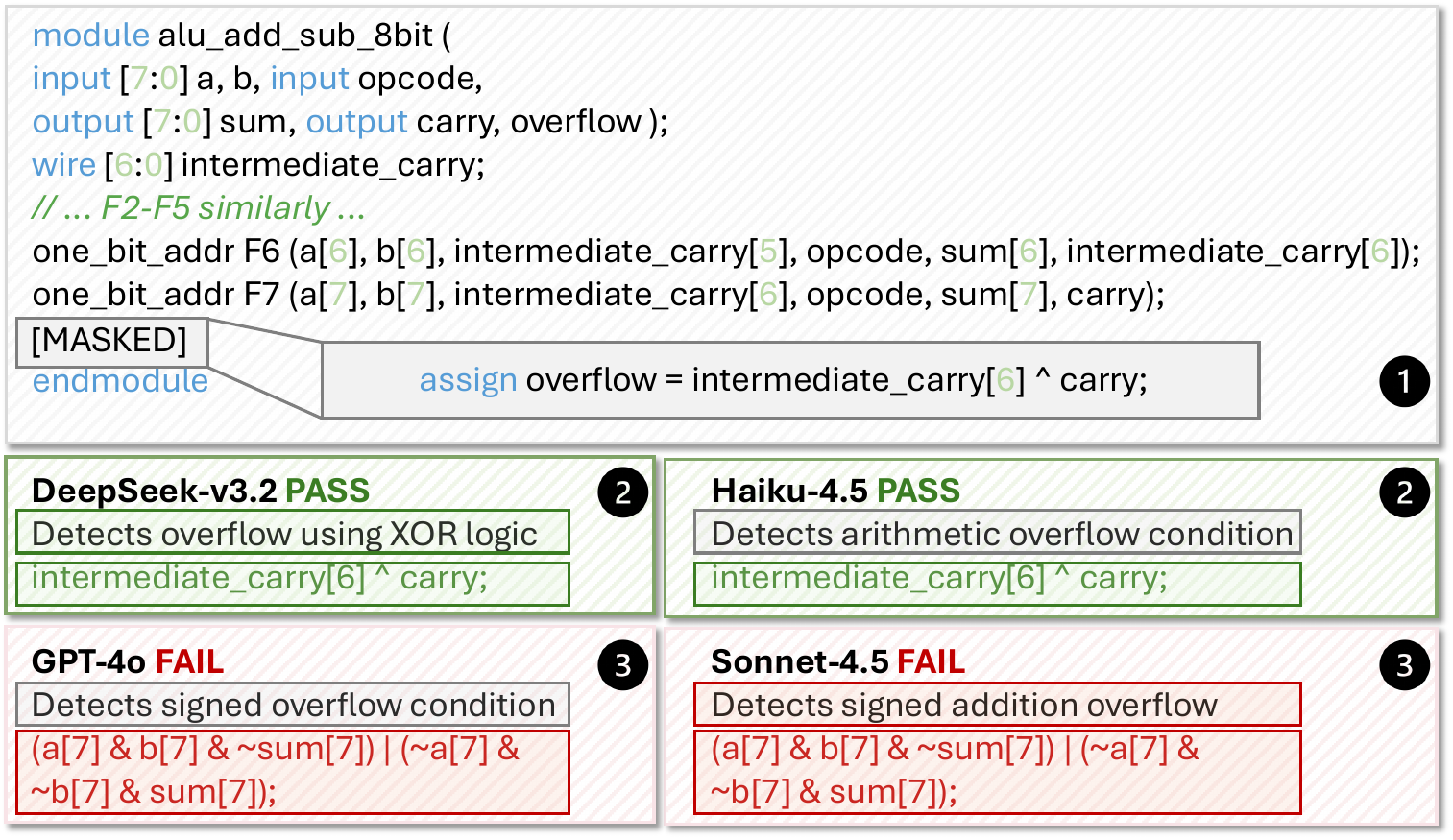}
    \caption{Task~2.3 case: models must infer the design-specific overflow convention rather than substitute a generic formula.}
    \label{fig:case-mask}
  \end{minipage}
\end{figureblock}

\subsection{Examples for issue-fixing}

\begin{figure}[htbp]
  \centering
  \includegraphics[width=0.98\linewidth]{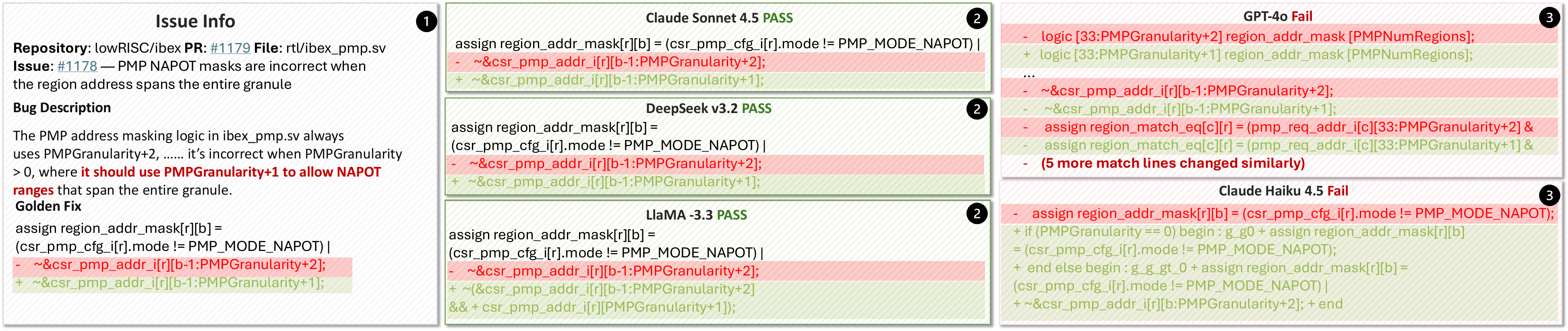}
  \caption{Task~3.2b case: the issue requires one precise bit-slice correction rather than a global textual replacement.}
  \label{fig:case-repo}
\end{figure}

\paragraph{Case 3: functional issue-fixing.}
Figure~\ref{fig:case-repo} uses lowRISC/ibex issue \#1178. The golden fix changes one slice bound in the PMP NAPOT address mask from \texttt{PMPGranularity+2} to \texttt{PMPGranularity+1}, while leaving nearby \texttt{+2} occurrences unchanged. Sonnet-4.5 and DeepSeek-v3.2 make the exact edit; LLaMA-3.3 produces a functionally equivalent rewrite. GPT-4o globally changes every \texttt{+2}, corrupting unrelated widths. Haiku-4.5 localizes the line but adds an unnecessary branch that fails formal equivalence. Passing repairs change at most one line, whereas these failed repairs change three or more.

\clearpage

\subsection{Examples on a project-level case: VGA/LCD Controller}
\label{app:vga-agent-comparison}

 Figure~\ref{fig:vga-generation-pair} shows the functional hierarchy and the five top-level testpoints for the case Example \texttt{opencores\_vga\_rgb\_v1}. Each testpoint is a complete simulation check. We provide the evaluation results on GPT-5.6 Codex CLI and DeepSeek OpenHands.

\textbf{Case 4: GPT-5.6 Sol with Codex CLI.} The agent read the public specification, module interfaces, and output contract, then wrote all 15 required RTL files. It ran the public checker, corrected an address-decoding warning, and exercised local pixel and frame-buffer simulations. The final submission passed the output-contract and project-compilation gates. In the frozen evaluation, pixels, buffer switching, and underrun passed; registers and timing failed, with testpoint pass rates of 3/5.

\textbf{Case 5: DeepSeek V3.2 with OpenHands.} The agent read the same inputs and wrote all 15 required RTL files. Its public checker reported an invalid register part-select and an invalid top-level output-port connection. The agent fixed the part-select, then reached its 32-iteration limit before repairing the port connection. The final output-contract gate passed, but elaboration failed because the generated \texttt{vga\_pgen.clut\_adr} output was connected to \verb|{cbsw, 8'd0}|, which contains a constant and cannot be driven. The generated RTL fails compilation, so all testpoints fail with compilation errors before functional simulations. This indicates the difficulty of project-level tasks for cutting-edge models.

\begin{figureblock}
  \begin{minipage}{\textwidth}
    \centering
    \includegraphics[width=\linewidth]{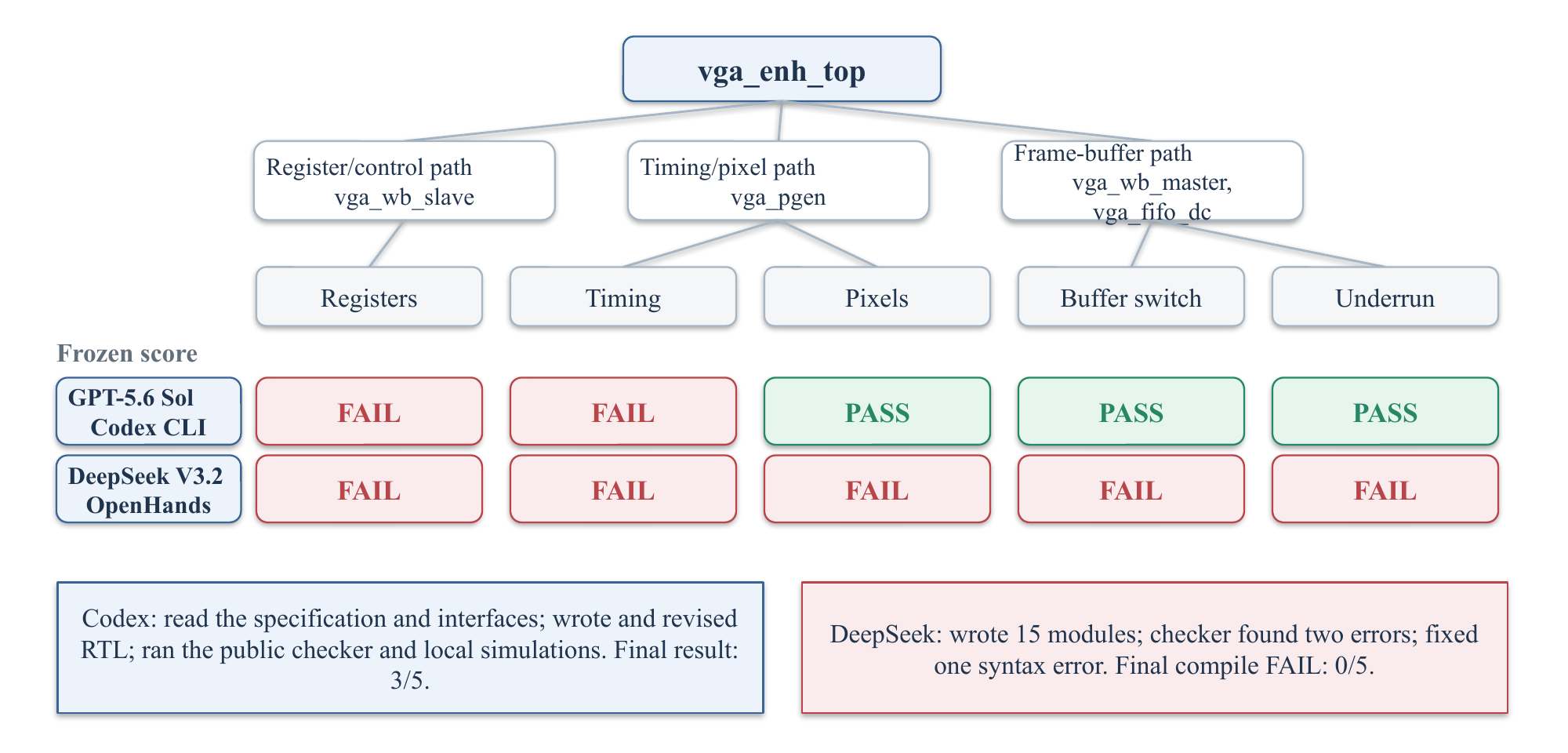}
    \caption{Two agent runs on the same VGA/LCD controller project.}
    \label{fig:vga-generation-pair}
  \end{minipage}
\end{figureblock}

\end{document}